\documentclass[11pt]{article}

\usepackage[final]{acl}

\usepackage{times}
\usepackage{latexsym}

\usepackage[T1]{fontenc}
\usepackage[utf8]{inputenc}

\usepackage{microtype}

\usepackage{inconsolata}

\usepackage{graphicx}

\usepackage{hyperref}
\usepackage{url}
\usepackage{amssymb}
\usepackage{subcaption}
\usepackage{booktabs}
\usepackage{multirow}
\usepackage{graphicx}
\usepackage{array}
\usepackage{enumitem}
\usepackage{float} 
\usepackage{amsmath}  
\usepackage{xcolor}
\title{When should I search more: Adaptive Complex Query Optimization\\ with Reinforcement Learning}

\author{
Wei Wen\footnote[1]{}\,$^{1}$,\,
Sihang Deng\footnote[1]{}\,$^{2}$\,
Tianjun Wei\thanks{{\footnotesize Equal Contribution.}}\footnote[2]{}\,$^{3}$,\,
Keyu Chen\,$^{1}$,\,
Ruizhi Qiao\footnote[2]{}\,$^{1}$,\,
Xing Sun\,$^{1}$\\
$^{1}$ Tencent Youtu Lab,
$^{2}$ The University of Hong Kong,
$^{3}$ Nanyang Technological University\\
\texttt{\{jawnrwen,yolochen,ruizhiqiao,winfredsun\}@tencent.com}\\\texttt{tjwei2-c@my.cityu.edu.hk}
}

\begin{document}
\maketitle
\begin{abstract}
  Query optimization is a crucial component for the efficacy of Retrieval-Augmented Generation (RAG) systems. While reinforcement learning (RL)-based agentic and reasoning methods have recently emerged as a promising direction on query optimization, most existing approaches focus on the expansion and abstraction of a single query. However, complex user queries are prevalent in real-world scenarios, often requiring multiple parallel and sequential search strategies to handle disambiguation and decomposition. Directly applying RL to these complex cases introduces significant hurdles. Determining the optimal number of sub-queries and effectively re-ranking and merging retrieved documents vastly expands the search space and complicates reward design, frequently leading to training instability.
  To address these challenges, we propose a novel RL framework called \underline{A}daptive \underline{C}omplex \underline{Q}uery \underline{O}ptimization (ACQO). Our framework is designed to adaptively determine when and how to expand the search process. It features two core components: an Adaptive Query Reformulation (AQR) module that dynamically decides when to decompose a query into multiple sub-queries, and a Rank-Score Fusion (RSF) module that ensures robust result aggregation and provides stable reward signals for the learning agent. To mitigate training instabilities, we adopt a Curriculum Reinforcement Learning (CRL) approach, which stabilizes the training process by progressively introducing more challenging queries through a two-stage strategy.
  Our comprehensive experiments demonstrate that ACQO achieves state-of-the-art performance on three complex query benchmarks, significantly outperforming established baselines. The framework also showcases improved computational efficiency and broad compatibility with different retrieval architectures, establishing it as a powerful and generalizable solution for next-generation RAG systems. %Code will be released upon acceptance.
  \end{abstract}
  
  \section{Introduction}
  Retrieval-Augmented Generation (RAG) has become a core paradigm in the LLM era because it grounds generation in external evidence, thereby improving factuality, recency, and attribution~\citep{huang2024survey,lewis2020retrieval}. Achieving these benefits in RAG hinges on obtaining high-quality retrieved evidence, which in turn depends on transforming a user’s natural-language question into a self-contained, retrieval-friendly query. This step is known as \textbf{Query Optimization (QO)}~\citep{yu2020few, vakulenko2021comparison, zhang2024adaptive}.
  
  Existing QO techniques primarily optimize a single query through expansion or abstraction~\citep{yu2020few, vakulenko2021comparison, zhang2024adaptive} in different approaches. Prompt-based approaches~\citep{azad2019query} leverage meticulously crafted instructions to guide the LLM in generating more effective search queries. For instance, a simple prompt might instruct the LLM to “rephrase the user's question to be more suitable for a search engine.”. Interactive-learning based methods~\citep{xu2024search, zhu2025convsearch, feng2023synergistic} go a step further by engaging in a feedback loop with the user or a simulated environment, allowing the model to refine its queries iteratively based on the quality of retrieved results. Pseudo-document generation techniques~\citep{wang2023query2doc,gao2023precise} transform the original query into a hypothetical, longer document that contains richer context, which can then be used to retrieve more relevant information from the knowledge base. More recently, agentic and reasoning-augmented reinforcement learning (RL) methods—valued for their reduced dependence on labeled supervision—have shown strong empirical gains \citep{singh2025agentic, zhu2025convsearch}. However, most of these solutions implicitly assume a one-to-one correspondence between a user query and an optimized query, which limits their coverage of complex information needs.
  
  In real-world RAG applications, complex queries are common and often require multiple parallel or sequential sub-queries, notably for disambiguation and decomposition \citep{song2024surveyqueryoptimizationlarge}.
  \vspace{-4pt}
  \begin{itemize}[leftmargin=10pt]
  % \item \textbf{Disambiguation queries}, such as a user asking, "\textit{Who is Bill Gates' father?}", require the system to identify and clarify ambiguous entities (e.g., distinguishing the Microsoft founder from his father). This may necessitate generating multiple parallel or sequential sub-queries to compare evidence.
  \item \textbf{Disambiguation queries}, such as a user asking, 
  ``\textit{When did Arsenal last win the FA Cup? [SEP] 2005 [SEP] What about them compared to Chelsea in league titles?}", 
  require the system to interpret multi-turn contexts and clarify entity references (e.g., linking "them" back to Arsenal while introducing Chelsea for comparison). 
  This may necessitate generating multiple parallel or sequential sub-queries to retrieve and contrast evidence.
  \item \textbf{Decomposition queries}, such as a user asking, ``\textit{What were the global shipments of iPhones in 2022 and 2023, respectively?}", require breaking down a multi-objective problem into independent sub-queries (e.g., "\textit{global iPhone shipments in 2022}" and "\textit{global iPhone shipments in 2023}"), retrieving results for each, and then synthesizing a final answer.
  \end{itemize}
  \vspace{-4pt}
  While some prior work has explored these problems~\citep{ammann2025question,perez2020unsupervised,liu2024ra}, applying reinforcement learning to such complex scenarios still presents a series of challenges: (1) deciding query number and depth (when to stop, whether to branch, how to merge); (2) performing multi-path retrieval and document aggregation across heterogeneous retrievers (sparse, dense, hybrid) with consistent, robust signals; and (3) coping with expanded search spaces and sparse/delayed rewards, which destabilize training.
  We argue that an effective QO system for complex queries should satisfy two goals:
  \vspace{-4pt}
  \begin{itemize}[leftmargin=10pt]
  \item \textbf{Adaptive query handling}: it should adaptively decide the number and depth of sub-queries and switch among disambiguation, decomposition and single-query expansion and abstraction.
  \item \textbf{Stability and integrability}: it should support an end-to-end pipeline (query reformulation → multi-retrieval → document re-ranking → answer generation), seamlessly integrate with sparse and dense retrieval backends, and incorporate stabilizing training mechanisms tailored to RL.
  \end{itemize}
  \vspace{-4pt}
  
  To meet these goals, in this paper we propose Adaptive Complex Query Optimization (ACQO), an RL framework that learns when and how to expand the search process and how to accumulate evidence robustly.
  First, we let LLM decide whether to trigger decomposition or disambiguation, producing a set of parallel or staged sub-queries based on query complexity and intent diversity.
  Then, we perform model-agnostic re-ranking and fusion by jointly exploiting rank positions and retrieval scores, enabling smooth integration with heterogeneous retrievers and providing stable intermediate signals for the RL agent. Finally, we introduce a Curriculum Reinforcement Learning (CRL) strategy with two stages: an initial phase for broad exploration over all samples to establish general policies, followed by a focused phase that emphasizes challenging cases. This curriculum mitigates reward sparsity and improves convergence stability across the spectrum of query complexities.
  In experiments, \textsc{ACQO} achieves state-of-the-art performance on widely used RAG benchmarks, including conversational query reformulation (TopiOCQA)~\citep{adlakha2022topiocqa} and multi-hop reasoning (HotpotQA)~\citep{yang2018hotpotqa}, with additional out-of-domain evaluation on MultiHop-RAG~\citep{tang2024multihoprag} demonstrating strong generalization capabilities. Notably, our lightweight components achieve performance comparable to approaches requiring specialized retrieval modifications or complex re-ranking architectures, while maintaining significantly lower computational overhead. Experimental results demonstrate substantial improvements over baseline methods in both quantitative metrics and qualitative analysis.
  The contributions of this work are as follows:
  \vspace{-4pt}
  \begin{itemize}[leftmargin=10pt]
  \item We propose \textsc{ACQO}, which unifies adaptive multi-query decision-making with robust evidence fusion in an end-to-end RL framework for complex queries.
  \item We introduce a universal re-ranking mechanism to combine rank positions and retrieval scores in a model-agnostic manner, improving stability and transferability across heterogeneous retrievers.
  % making it broadly applicable across heterogeneous retrievers.
  % \item We introduce a universal re-ranking mechanism to combine rank positions and retrieval scores in a model-agnostic manner, improving stability and transfer across heterogeneous retrievers.
  \item Through extensive experiments on benchmark datasets, we demonstrate that \textsc{ACQO} significantly outperforms existing methods while maintaining computational efficiency, establishing its superiority for complex query processing in RAG systems.
  \end{itemize}
  \vspace{-4pt}
  
  \section{What makes queries complex in real-world RAG scenarios?}
  In this section, we conduct a systematic analysis of query complexity patterns in real-world RAG benchmark. By examining the inherent characteristics of queries across different datasets, we identify the key challenges that motivate our ACQO framework design.

  \begin{figure}[htbp!]  
    \centering
    \includegraphics[width=0.95\linewidth]{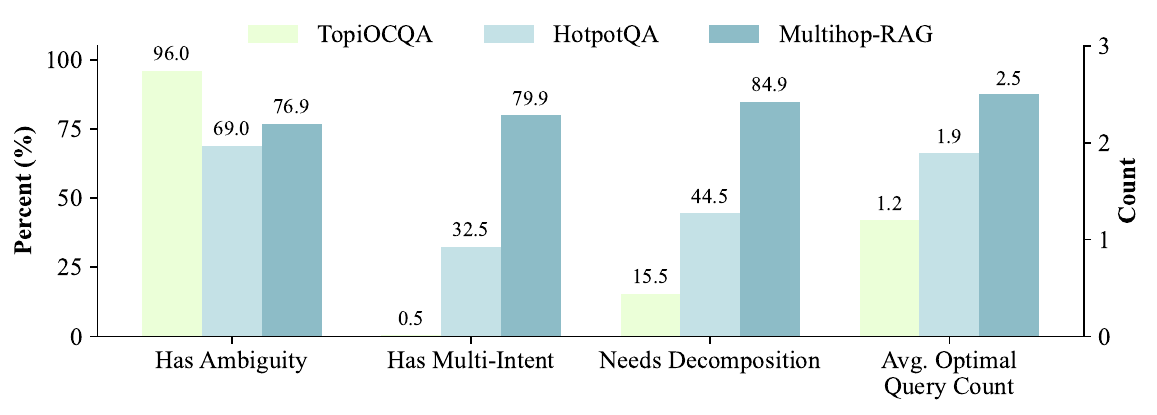}
    \caption{Distribution of query complexity.}  
    \label{fig:complexity_distribution}  
  \end{figure}

  \begin{figure}[htbp!]
    \centering
    \includegraphics[width=0.7\linewidth]{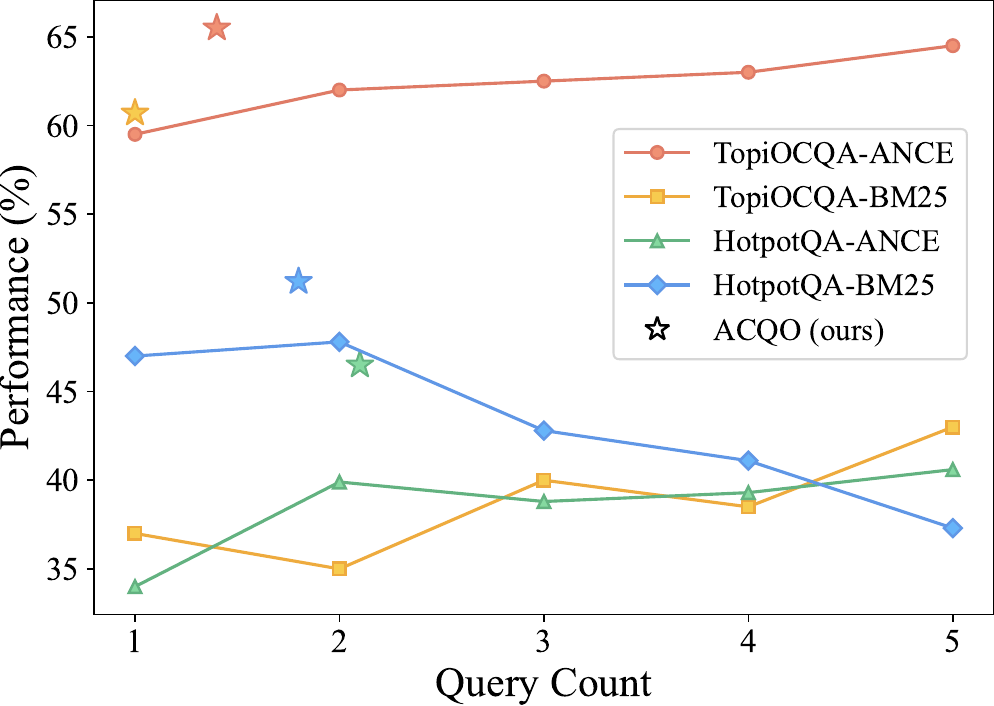}
    \caption{Performance with different query counts.}  % 独立标题
    \label{fig:performance_querycount}  
  \end{figure}

  \subsection{Query Complexity Analysis Framework}
  We analyze three representative RAG benchmarks:
  %, each emphasizing a distinct reasoning paradigm:
  \textsc{TopiOCQA} for multi-turn conversational QA, \textsc{HotpotQA} for multi-hop factual reasoning, and \textsc{MultiHop-RAG} for real-world multi-hop retrieval. 
  For each query, we conduct a structured analysis using the following criteria:
  \begin{itemize}[leftmargin=10pt]
  \item \textbf{Ambiguity Detection}: Flag ambiguous entities or references that need disambiguation.
  \item \textbf{Multi-Intent Analysis}: Identify distinct intents embedded in the query.
  \item \textbf{Decomposition Assessment}: Judge whether decomposition improves answerability.
  \item \textbf{Optimal Granularity}: Identify the minimum number of sub-queries from the generated set that yields optimal retrieval metrics.
  \end{itemize}
  We analyze 200 representative queries from each dataset, focusing on understanding the distribution and characteristics of complex queries in real-world scenarios.
  
  \vspace{-3pt}
  \subsection{Dataset Analysis: Prevalence of Complex Queries}
  
  \begin{table}[t]
  \scriptsize
  \centering
  \scriptsize
  \setlength\tabcolsep{3.5pt}
  \renewcommand{\arraystretch}{0.6}
  \begin{tabular}{l|cc|cc|cc|cc}
  \toprule
  \multirow{3}{*}{\textbf{Method}} & \multicolumn{4}{c|}{\textbf{TopiOCQA (Recall@10)}} & \multicolumn{4}{c}{\textbf{HotpotQA (MAP@10)}} \\
  \cmidrule(lr){2-5} \cmidrule(lr){6-9}
  & \multicolumn{2}{c|}{ANCE} & \multicolumn{2}{c|}{BM25} & \multicolumn{2}{c|}{ANCE} & \multicolumn{2}{c}{BM25} \\
  \cmidrule(lr){2-3} \cmidrule(lr){4-5} \cmidrule(lr){6-7} \cmidrule(lr){8-9}
  & Easy & Hard & Easy & Hard & Easy & Hard & Easy & Hard \\
  \midrule
  Prompt-based & 59.4 & 52.6 & 34.3 & 45.1 & 36.8 & 31.2 & 50.1 & 40.5 \\
  SFT & 56.2 & 54.8 & 33.1 & 38.7 & 44.7 & 33.5 & 45.2 & 43.7 \\
  Vanilla RL & 63.9 & 54.8 & 58.5 & 61.2 & 42.3 & 38.2 & 50.4 & 46.0 \\
  \midrule
  ACQO (ours) & \textbf{66.2} & \textbf{58.0} & \textbf{60.3} & \textbf{64.5} & \textbf{50.4} & \textbf{41.5} & \textbf{53.1} & \textbf{48.3} \\
  \bottomrule
  \end{tabular}
  \caption{Performance comparison on easy vs. hard query subsets across datasets and retrievers (\%).}
  \label{tab:easy_hard_analysis}
  \vspace{-8pt}
  \end{table}
  
  % \begin{table}[h]
  % \centering
  % \caption{Query Complexity Distribution Across Datasets}
  % \label{tab:complexity_distribution}
  % \begin{tabular}{lcccc}
  % \toprule
  % \textbf{Characteristic} & \textbf{TopiOCQA} & \textbf{HotpotQA} & \textbf{MultiHop-RAG} & \textbf{Average} \\
  % \midrule
  % Has Ambiguity & 96.0\% & 69.0\% & 76.9\% & 80.6\% \\
  % Has Multi-Intent & 0.5\% & 32.5\% & 79.9\% & 37.6\% \\
  % Needs Decomposition & 15.5\% & 44.5\% & 84.9\% & 48.3\% \\
  % Avg. Optimal Query Count & 1.2 & 1.5 & 2.5 & - \\
  % \bottomrule
  % \end{tabular}
  % \end{table}
  Our structured analysis reveals significant complexity patterns across the three toy datasets, with Figure~\ref{fig:complexity_distribution} illustrating the distribution of query characteristics. Specifically, a substantial proportion of queries
  % in real-world RAG scenarios 
  are complex: on average, 48.3\% require decomposition, and 37.6\% exhibit multiple intents. Moreover, the optimal number of sub-queries varies across domains (1.2–2.5 on average), indicating that decomposition strategies must be context-sensitive rather than one-size-fits-all. 
  
  \vspace{-3pt}
  \subsection{Why Current Methods Struggle with Complex Queries}
  We evaluate representative query optimization approaches across different paradigms: prompt-based optimization using \textit{DeepSeek-V3.1}~\citep{deepseekai2024deepseekv3technicalreport} with decomposition prompts, supervised fine-tuning (SFT) via \textit{Qwen2.5-3B}~\citep{qwen2.5} query rewriter, and vanilla reinforcement learning (REINFORCE with sparse rewards) also based on \textit{Qwen2.5-3B}. %Our analysis reveals four fundamental limitations that motivate the ACQO framework.
  
  %The performance analysis in Table~\ref{tab:easy_hard_analysis} reveals critical limitations of existing approaches when handling complex queries. Current methods exhibit substantial performance variations between easy and hard queries, with prompt-based approaches showing dramatic drops of up to 41.4\% on TopiOCQA (38.4\% to 22.5\% with ANCE), indicating a lack of adaptive mechanisms for varying query complexity. Moreover, optimal approaches vary significantly across retrieval systems—vanilla RL excels with BM25 (62.2\%) but degrades with ANCE (54.8\%) on hard queries, while SFT maintains more consistent cross-retriever performance, suggesting that current methods fail to adapt their optimization strategies to different retrieval characteristics. Figure~\ref{fig:performance_querycount} further illustrates the suboptimal nature of fixed decomposition strategies, which also suffer from dual limitations in both efficiency and effectiveness. 
  %These inconsistent performance patterns across query types and retrieval systems highlight the absence of principled approaches for systematic query optimization, as current methods employ fixed strategies regardless of query characteristics or system properties. These empirical findings reveal three critical gaps that motivate our ACQO framework: the need for adaptive complexity recognition, retriever-aware optimization, and integration for decomposed queries.
  
  The performance analysis in Table~\ref{tab:easy_hard_analysis} reveals critical limitations of existing approaches when handling complex queries. Current methods exhibit substantial performance variations between easy and hard queries, with SFT approaches showing dramatic drops of up to 11.2\% on HotpotQA (44.7\% to 33.5\% with ANCE). Moreover, optimal approaches vary significantly across retrieval systems—vanilla RL excels with BM25 (61.2\%) but degrades with ANCE (54.8\%) on hard queries.%, while SFT maintains more consistent performance. 
  Figure~\ref{fig:performance_querycount} further demonstrates that fixed decomposition strategies suffer from dual limitations in both efficiency and effectiveness. These inconsistent patterns highlight the absence of principled approaches for systematic query optimization, revealing three critical gaps: adaptive complexity recognition, retriever-aware optimization, and effective integration for decomposed queries.
  
  %%检索器相关，如何表达
  
  % \begin{table}[h]
  % \centering
  % \caption{Performance Comparison Across Different Retrievers (Recall@10)}
  % \label{tab:method_comparison}
  % \begin{tabular}{lcc}
  % \toprule
  % \textbf{Method} & \textbf{BM25} & \textbf{ANCE}  \\
  % \midrule
  % \multicolumn{5}{l}{\textit{Simple Queries}} \\
  % Prompt-based &  &   \\
  % SFT &  &    \\
  % Vanilla RL &  &   \\
  % \midrule
  % \multicolumn{5}{l}{\textit{Complex Queries}} \\
  % Prompt-based &  &  \\
  % SFT &  &   \\
  % Vanilla RL &  &   \\
  % \midrule
  % Performance Drop &  &    \\
  % \bottomrule
  % \end{tabular}
  % \end{table}

  % \subsubsection{Detailed Analysis Examples}

  \section{\underline{A}daptive \underline{C}omplex \underline{Q}uery \underline{O}ptimization}
  \label{gen_inst}
  
  \subsection{Task Formulation}
  In traditional Query Optimization (QO) pipeline, the task is defined as refining the query to retrieve the golden document(s) relevant to the user’s current query and conversational context (if any) from a large collection of documents. 
  Formally, given the current query $q^{(t)}$ ($t \geq 1$) and its historical context $C^{(t-1)} = \{ (q_i, a_i) \}_{i=1}^{(t-1)} \quad (\text{if } t \geq 2)$, where $t$ denotes the current turn number, a query optimization model ${\Theta}$ generates a de-contextualized query $\hat{q}^{(t)}$c. $\hat{q}$ ($(t)$ is omitted for simplicity) is subsequently input into a retrieval system, which returns a ranked list of the top-$k$ documents from the collection $\mathcal{P}$. We denote this ranked set as $\mathcal{R}_k(\hat{q}) = \{ p_1, p_2, \dots, p_k \}, \quad \mathcal{R}_k(\hat{q}) \subseteq \mathcal{P}$, where $p_i$ represents the document ranked at position $i$. Let $\mathcal{P}^* \subseteq \mathcal{P}$ denote the set of golden documents corresponding to $\hat{q}$. The objective of QO is (1) to maximize the probability that at least one golden document in $\mathcal{P}^*$ appears in $\mathcal{R}_k(\hat{q})$; and (2) to minimize the ranking positions of the golden documents within $\mathcal{R}_k(\hat{q})$.  
  
  In our work, we extend this formulation by considering the disambiguation and decomposition scenarios, where an optimized query set $\hat{\mathcal{Q}}_{q}$ will be generated. 
  Each sub-query $\hat{q}_q \in \hat{\mathcal{Q}_q}$ retrieves its own top-$k$ documents $\mathcal{R}_k(\hat{q}_q)$, and these candidates are subsequently combined and re-ranked to produce the final top-$k$ documents, denoted as $\mathcal{R}_k(\hat{\mathcal{Q}_q})$. This design enhances both the coverage and ranking quality of golden documents.
  
  \subsection{Overall Framework}
  % The proposed AQCO is a RL framework that learned when and how to expand the search process and how to accumulate evidence robustly.
  % It is applicable to both multi-hop queries and multi-turn conversational queries, where effective reformulation is essential for retrieving relevant documents.
  % % 这里应该改一下，引出为什么需要re-ranker
  
\begin{figure}[t]
    \centering
    % 这里是图片内容（或占位框）
    \includegraphics[width=\linewidth]{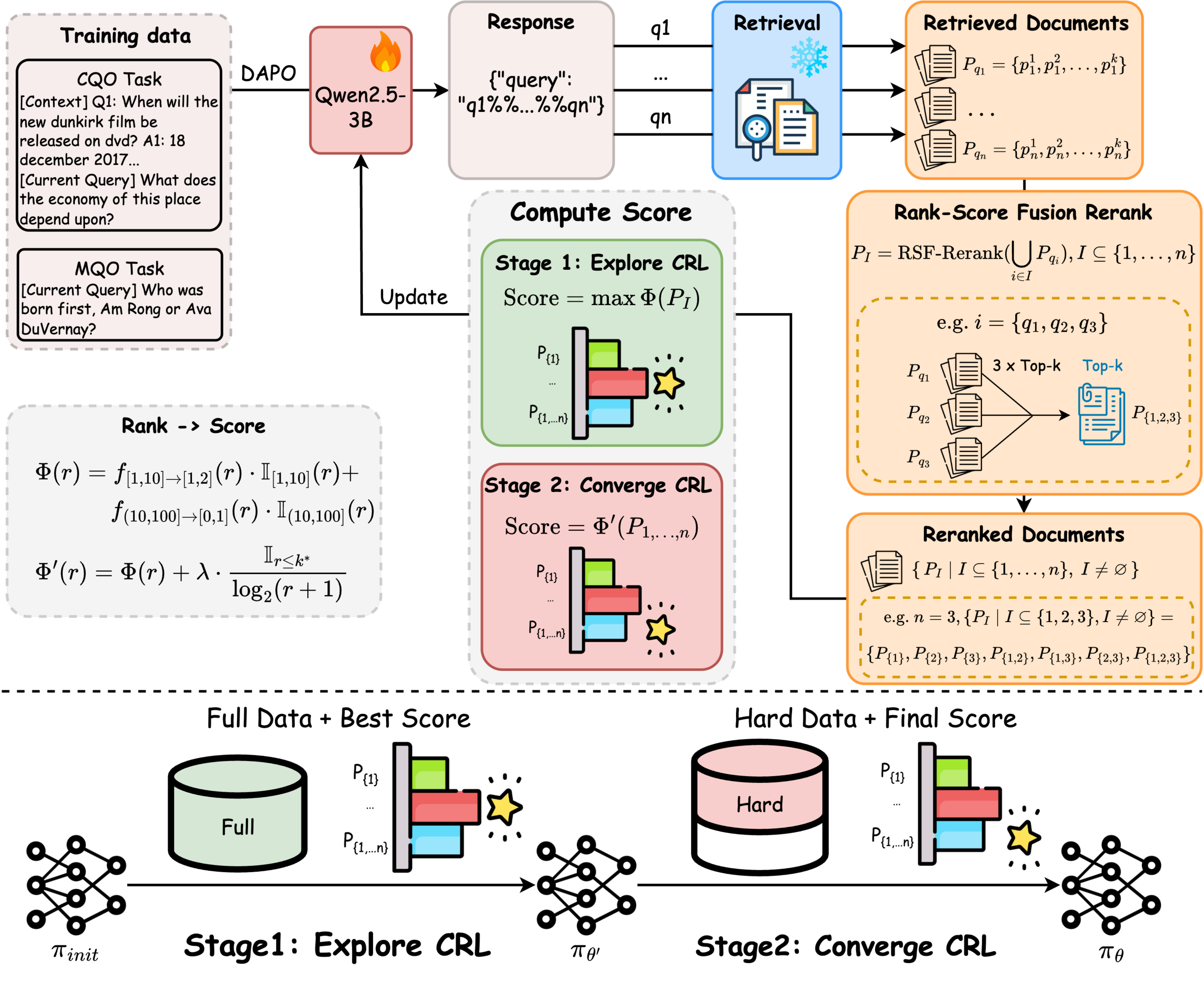}
    \caption{Overview of ACQO. ACQO employs two-stage curriculum reinforcement learning to adaptively optimize complex queries and integrate multi-retrieval results via Rank-Score Fusion.}  % 图片标题
    \label{fig:Overview}  % 为图片定义标签（关键步骤）
  \end{figure}

  As illustrated in \autoref{fig:Overview}, ACQO proceeds in two curriculum reinforcement learning (CRL) stages: 
  (1) \emph{Explore CRL}, which promotes broad exploration and early stabilization; and 
  (2) \emph{Converge CRL}, which emphasizes precision and convergence on harder cases. 
  
  The core idea is to integrate query optimization with CRL in a fully self-directed manner. 
  Without external supervision or intervention, the model learns to adaptively converge to suitable query numbers and optimization strategies across heterogeneous retrieval systems. 
  In the following, we first introduce our re-ranker design, which consolidates multiple retrieval lists produced from the query set, and then detail the two-stage CRL procedure.
  %we introduce our re-ranker design—responsible for consolidating multiple retrieval lists from query set—before detailing the two-stage CRL procedure.
  
  \subsection{Re-ranker Design}
  \paragraph{Method.} Inspired by Reciprocal Rank Fusion (RRF), we propose a new method named \textbf{Rank-Score Fusion (RSF)} to address two key limitations of RRF: it only considers rank positions while ignoring absolute retrieval scores, and it cannot properly handle cases where documents obtain identical ranks across multiple lists.
  
  In RSF, each sub-query returns a ranked list of candidate documents, where each document is associated with a retrieval(e.g., ANCE) score and a rank position. For a given document $p$, we collect its appearances across all $M$ sub-queries into a set
  $\{(s_j, r_j)\}_{j=1}^M$ ,where $(s_j, r_j)$ denotes the score and rank of document $p$ in the $j$-th sub-query. 
  We then compute two aggregated quantities for $p$: 
  \begin{equation}
  \small
  P(p) = \frac{1}{\sum_{j=1}^{M} \frac{1}{r_j}}, \quad 
  S(p) = \max_{j=1,\dots,M} s_j.
  \end{equation}
  Here, $P(p)$ reflects the combined influence of rank positions (relative values), while $S(p)$ captures the strongest absolute score observed for document $p$. 
  We therefore perform lexicographical sorting with $P(p)$ as the primary key (ascending order: lower rank indicates better consensus) and $S(p)$ as the secondary key (descending order: higher score indicates stronger evidence). Formally, candidate documents are re-ranked according to:
  \begin{equation}
  \small
  \begin{aligned}
  \mathcal{R}_k = \text{Top-}k \big( &\text{sort}\{ (p, P(p), S(p)) \mid \\ &p \in \mathcal{R}_k(q_1) \cup \cdots \cup \mathcal{R}_k(q_M) \}\big),
  \end{aligned}
  \end{equation}
  where the sorting key is $(P(p), -S(p))$ in ascending order. This encodes a hierarchical preference: "Trust rank consensus first; use scores only to break ties among similarly-ranked documents." %Critically, since all sub-queries within a query instance use the same retriever, $S(p)$ values are naturally comparable without normalization—avoiding the instability issues of weighted-sum approaches that require score normalization~\citep{wu2020variance}.

  % \paragraph{Advantages.}
  % Our RSF method inherits the simplicity and efficiency of RRF while further extending its capability by incorporating score information. Specifically, RSF offers several advantages:
  % \begin{enumerate}[leftmargin=10pt]
  %     \item \textbf{No additional latency and compatibility with re-rank models:} similar to RRF, RSF introduces no extra inference delay and can be seamlessly combined with neural re-rankers at inference time.
  %     \item \textbf{Adaptability to heterogeneous retrieval systems:} RSF, like RRF, can be directly applied to both sparse (e.g., BM25) and dense (e.g., ANCE) retrieval methods, and is compatible with different index structures such as flat indexes (where higher scores indicate better similarity) and HNSW graphs (where lower scores indicate better similarity).
  %     \item \textbf{Joint consideration of rank and score:} beyond RRF’s rank-only design, RSF leverages both relative positions and absolute similarity values, achieving a more robust and balanced re-ranking, while naturally resolving the ambiguity when documents share identical ranks.
  % \end{enumerate}
  
  \paragraph{Advantages.}
  Our RSF method inherits RRF's simplicity and efficiency while extending its capability through score integration. RSF offers three key advantages: 
  (1) \textbf{Zero latency overhead}: introduces no inference delay and seamlessly integrates with neural re-rankers.
  (2) \textbf{Universal compatibility}: directly applicable to both sparse (e.g., BM25) and dense (e.g., ANCE) retrievers across different index structures.
  (3) \textbf{Enhanced robustness}: leverages both rank positions and absolute scores for more balanced re-ranking while resolving rank ambiguities.

  \subsection{Curriculum Reinforcement Learning}
  \label{subsec:CRL}
  \subsubsection{Base Reward Function}
  
  We build upon the Rank-Incentive Reward Shaping (RIRS) framework proposed in ConvSearch-R1~\citep{zhu2025convsearch}, which provides dense rank-based reward signals and alleviates the sparsity of traditional metrics such as NDCG and MRR. 
  Here, the rank $r$ is defined as the position assigned to a document in the re-ranked list $\mathcal{R}_k$ from our RSF module. 
  The base rank-to-score mapping employs a continuous piecewise linear transformation:
  \begin{equation}
  \small
  \begin{aligned}
  \Phi(r) =&
  f_{[1,10] \rightarrow [1,2]}(r) \cdot \mathbb{I}_{[1,10]}(r) \\ +& f_{(10,100] \rightarrow [0,1)}(r) \cdot \mathbb{I}_{(10,100]}(r),
  \end{aligned}
  \end{equation}
  where $f_{A \rightarrow B}$ represents a linear mapping function from interval $A$ to interval $B$, $\mathbb{I}_A(r)$ is the indicator function that equals 1 when $r \in A$ and 0 otherwise, and $r$ is the rank variable.
  
  To accommodate multiple relevant documents, we employ a weighted aggregation score emphasizes the most promising retrieval results. Suppose the rank of $n$ retrieved relevant documents in ranked set $\mathcal{R}$ are $r_1, r_2, ..., r_n$ respectively, the $r_i$ score is defined as:
  \begin{equation}
  \small
  s(r_i) =  \eta^i \cdot \Phi(r_i),
  \end{equation}
  where $\eta$ is the decay coefficient.  This generalization retains the dense reward structure of RIRS while providing additional flexibility to adapt the weighting scheme for different retrieval scenarios. 
  
  Taking the format correctness into the consideration, the complete reward score is defined as:
  \begin{equation}
  \small
  S(\mathcal{R}) = \sum_{i=1}^{n}s(r_i) \cdot \mathbb{I}_{format} + \delta \cdot (1 - \mathbb{I}_{format})
  \end{equation}
  where $\mathbb{I}_{format}$ serves as the format compliance gate, and $\delta < 0$ represents the non-compliance penalty coefficient.
  % Finally, following ConvSearch-R1~\citep{zhu2025convsearch}, we incorporate a format compliance indicator $\phi \in \{0,1\}$, with a small penalty $\delta<0$ for non-compliant queries:
  % \begin{equation}
  
  % \mathcal{R} =
  % \begin{cases}
  % S, & \phi = 1, \\begin{equation}
  
  % \delta, & \phi = 0.
  % \end{cases}
  % \normalsize

  \subsubsection{Stage I: Explore-Oriented CRL}
  
  \paragraph{Data Curriculum.} 
  In the exploration stage, we employ the full training dataset without filtering. 
  This ensures that the model is exposed to both easy and hard cases, providing sufficient diversity to stabilize early training and improve robustness. 
  By leveraging the entire dataset, the model can better explore the space of optimization without being biased toward specific difficulty levels.
  
  \paragraph{Reward Design.} 
  Building upon the base reward function, Stage I encourages exploration by reinforcing the \emph{combination of the best-performed sub-queries}.  
  Suppose $\hat{\mathcal{Q}}$ is the set of optimized sub-queries, and for any non-empty subset $\hat{\mathcal{Q}}'$ in the power set of $\hat{\mathcal{Q}}$, denoted as $\mathcal{P}(\hat{\mathcal{Q}})$, we compute its the stage-specific reward as:
  \begin{equation}
  \small
  G^{(I)}(\hat{\mathcal{Q}}) = 
  \max_{\hat{\mathcal{Q}}' \in \mathcal{P}(\hat{\mathcal{Q}}) \setminus \varnothing} 
  S(\mathcal{R}_k(\hat{\mathcal{Q}}')).
  \end{equation}
  This design allows the model to explore diverse decomposition strategies and ensures that promising sub-queries are strongly reinforced, even in the early stage when the model is not yet stable.
  
  \begin{table*}[t]
    \centering
    \scriptsize
    \setlength{\tabcolsep}{4pt}
    \renewcommand{\arraystretch}{0.6}\
  
    \begin{tabular}{l cc | cccc | cccc}
    \toprule
    \multirow{2}{*}{\textbf{Method}} & \multirow{2}{*}{\textbf{NS}} & \multirow{2}{*}{\textbf{NCoT}} & \multicolumn{4}{c|}{\textbf{Sparse(BM25)}} & \multicolumn{4}{c}{\textbf{Dense(ANCE)}} \\
    \cmidrule(lr){4-7} \cmidrule(lr){8-11}
     &  &  & MRR@3 & NDCG@3 & R@10 & R@100 & MRR@3 & NDCG@3 & R@10 & R@100 \\
    \midrule
    DeepSeek-V3.1                & - & - & 15.5 & 17.0 & 36.7 & 65.3 & 28.4 & 30.8 & 56.3 & 77.8 \\
    vanilla RL \textit{\textcolor{gray}{(Qwen2.5-3B)}}                & - & - & 31.2 & 36.1 & 60.8 & 82.5 & 34.5 & 38.3 & 62.1 & 81.1 \\
    IterCQR \textit{\textcolor{gray}{(T5-base)}}              & $\times$ & \checkmark & 16.5 & 14.9 & 29.3 & 54.1 & 26.3 & 25.1 & 42.6 & 62.0 \\
    ADACQR \textit{\textcolor{gray}{(T5-base+LLaMA7B)}}       & $\times$ & \checkmark & 28.3 & 26.5 & 48.9 & 71.2 & 38.5 & 37.6 & 58.4 & 75.0 \\
    LLM4CS-RAR \textit{\textcolor{gray}{(ChatGPT)}}           & \checkmark & $\times$ & 27.9 & 26.4 & 48.4 & 71.1 & 35.4 & 34.4 & 55.2 & 72.2 \\
    CHIQ-Fusion \textit{\textcolor{gray}{(T5-base+LLaMA2-7B)}}& $\times$ & \checkmark & 25.6 & 23.5 & 44.7 & --   & 38.0 & 37.0 & 61.6 & -- \\
    RETPO \textit{\textcolor{gray}{((LLaMA2-7B)}}             & $\times$ & \checkmark & 28.3 & 26.5 & 48.3 & 73.1 & 32.2 & 31.1 & 51.6 & 69.5 \\
    AdaQR \textit{\textcolor{gray}{(T5-base)}}                & $\times$ & \checkmark & 20.3 & 18.0 & 37.1 & 66.2 & \underline{38.1} & 36.6 & 61.3 & 79.9 \\
    ConvSearch-R1 \textit{\textcolor{gray}{(Qwen2.5-3B)}}     & $\times$ & $\times$ & \textbf{37.8} & \underline{36.2} & \underline{59.6} & \underline{80.1} & \textbf{50.5} & \textbf{50.1} & \textbf{72.0} & \textbf{86.3} \\
    \midrule
    ACQO \textit{\textcolor{gray}{(ours, Qwen2.5-3B)}}        & \checkmark & \checkmark & \underline{34.9} & \textbf{37.7} & \textbf{62.6} & \textbf{83.2} & 36.6 & \underline{39.4} & \underline{65.6} & \underline{85.1} \\
    \bottomrule
    \end{tabular}
    \caption{Retrieval performance comparison on TopiOCQA (\%). \textbf{NS} denotes training without rewrite supervised data, and \textbf{NCoT} denotes training without chain-of-thought reasoning.}
    \label{tab:topiocqa_results}
  
    \end{table*}
    
    \vspace{-4pt}

  \subsubsection{Stage II: Converge-Oriented CRL}
  \label{subsec:stage2crl}
  \paragraph{Data Curriculum.} 
  In the convergence stage, we refine the training distribution by focusing on the tougher cases. Rather than arbitrary filtering, we identify the optimal learning frontier by analyzing the performance distribution of Stage I models.
  
  Formally, let $\mathcal{Q}_{train}$ denote the full training query set. We define the \emph{learning complexity score} for each input query $q$ as:
  \begin{equation}
  \small
  \tau(q) = \frac{1}{K}\sum_{k=1}^{K}  G^{(I)}(\mathcal{\hat{Q}}_q^{(k)}),
  \end{equation}
  where %$\hat{Q}_q^{(k)}$ denotes a rollout optimized sub-query set, and 
  $K$ denotes the number of rollouts. The convergence curriculum $\mathcal{Q}_{conv}$ is constructed by retaining samples within \emph{optimal challenge zone}:
  \begin{equation}
  \small
  \mathcal{Q}_{conv} = \{q \in \mathcal{Q}_{train} : \tau(x_i) \leq \tau_{thres}\}
  \end{equation}
  where $\tau_{thres}$ is the theoretical boundary indicating when retrieval performance is sufficient to continue learning without destabilizing optimization.
  %corresponds to the theoretical boundary where retrieval performance indicates sufficient complexity for continued learning without overwhelming the optimization process. 
  % Here we set $\tau_{thres}=\frac{5}{3}$.

  This principled approach ensures that the model focuses on samples that are neither trivially easy (already mastered) nor prohibitively difficult (leading to sparse learning signals), thereby maximizing learning efficiency in the convergence phase.
  
  % Specifically, we filter the data by discarding instances where the Stage I model retrieves relevant documents within rank $\leq 3$. 
  %Specifically, we discard samples where the Stage I model (with 8 rollouts) yields a score $> 1.67$ (which corresponds to rank $\leq 3$).
  %This filtered subset emphasizes difficult queries and forces the model to concentrate on fine-grained improvements, enhancing efficiency and stability in later training.
  
  \paragraph{Reward Design.} 
  Stage II transitions from exploratory reward maximization to precision-focused optimization via a reward architecture that emphasizes ranking quality over quantity exploration.  The Stage II reward function directly evaluates the complete sub-query ensemble:
  %In contrast to Stage I, Stage II directly evaluates the entire sub-query set with the base reward function:
  \begin{equation}
  \small
  G^{(II)}(\hat{\mathcal{Q}}) =  
  S(\mathcal{R}_k(\hat{\mathcal{Q}})).
  \end{equation}
  %To further emphasize the importance of top-ranked results, we modify the rank-to-score function with explicit bonuses:
  
  To address the inherent challenge of sparse positive signals in top-ranked positions, we introduce a \emph{logarithmic precision weighting} mechanism, inspired by NDCG’s theoretical foundation, which reflects the information-theoretic principle that higher-ranked results contribute exponentially more to user satisfaction, which is defined as:
  \begin{equation}
  \small
  \Phi'(r) = \Phi(r) + \lambda \cdot  \frac{\mathbb{I}_{r \leq k^*}}{\log_2(r+1)},
  \end{equation}
  where $\lambda > 0$ is a precision amplification parameter, and $k^*$ represents the critical ranking threshold, $\mathbb{I}_{r \leq k^*}$ is the indicator function ensuring bonuses apply only to top-tier results.
  
  This bonus-based design provides stronger incentives for exact top placements while still leveraging the smooth decay of $ \Phi(r)$ for other positions. 
The model then gradually shifts from broad exploration in Stage I to precise convergence in Stage II.

  \section{Experiments}
  %We conduct comprehensive experiments to evaluate the effectiveness of our ACQO framework across multiple challenging datasets. Our experimental design covers multiple complex query optimization scenarios, including query disambiguation and query decomposition, with additional out-of-domain evaluation to assess generalization capabilities.
  \subsection{Experiments Setup}
  \paragraph{Datasets.}
  We train and evaluate our model on three representative benchmarks that cover both \textbf{multi-turn} conversational query optimization, which primarily focus on query \textbf{disambiguation}, and \textbf{multi-hop} query optimization task focusing on query \textbf{decomposition}. 
  For disambiguation task, we use TopiOCQA~\citep{adlakha2022topiocqa}, a challenging open-domain conversational QA dataset with topic shifts.
  For decomposition task, we adopt HotpotQA~\citep{yang2018hotpotqa} and evaluate generalization on MultiHop-RAG~\citep{tang2024multihoprag}, a RAG-focused multi-hop retrieval benchmark.
  %Dataset details are in Appendix~\ref{appendix:dataset}.
  
  %For CQR task, we select \textbf{TopiOCQA}~\citep{adlakha2022topiocqa}, an open-domain conversational dataset with topic switches, making it one of the most challenging datasets in this domain.  
  %For MQR task, we use \textbf{HotpotQA}~\citep{yang2018hotpotqa}, a widely used multi-hop question answering benchmark where each question is associated with 2 supporting documents. We also test the retrieval performance of model trained on HotpotQA on \textbf{Multihop-RAG}~\citep{tang2024multihoprag}, a benchmark specifically designed to evaluate multi-hop retrieval for RAG systems. More details of these datasets are provided in Appendix~\ref{appendix:dataset}.

  % \paragraph{Retrieval Systems.}
  % We evaluated the performance of model under both sparse and dense retrievers. 
  % For TopiOCQA and HotpotQA, we select \textbf{BM25} as the sparse retriever and ANCE as the dense retriever, where \textbf{ANCE}~\citep{xiong2020ance} is trained on MS-MARCO~\citep{bajaj2016msmarco} document retrieval tasks.
  % For MultiHop-RAG, we use \textbf{bge-large-en-v1.5}~\citep{xiao2024bge-large-en-v1.5} and \textbf{llm-embedder}~\citep{zhang2023llm-embedder} as the retrievers.
  
  \paragraph{Baselines.}
  
  We compare against three categories of prior work. For single query optimization reformulation and abstarction, we include \textit{IterCQR}~\citep{jang2023itercqr}, \textit{ADACQR}~\citep{lai2024adacqr}, and \textit{ConvSearch-R1}~\citep{zhu2025convsearch}. For query optimization with expansion, we evaluatedd \textit{LLM4CS-RAR}~\citep{mao2023large}, \textit{CHIQ-Fusion}~\citep{mo2024chiq}, \textit{RETPO}~\citep{yoon2025ask}, and \textit{AdaQR}~\citep{zhang2024adaptive}. For the complex query optimization setting, as there are no dedicated methods, we construct few-shot prompting baselines by adapting the above methods. We report post optimization retrieval performance after applying each baseline’s optimization procedure.

  The details regarding retriever, implementation and evaluation metrics are provided in Appendix~\ref{App:expdetails}.

  % baselines, implementation, evaluation metrics..

    \begin{table}[t]
    \tiny
    \centering
    \renewcommand{\arraystretch}{0.5}
    \setlength{\tabcolsep}{1.25pt}
    \begin{tabular}{@{}lcccccc@{}}
      \toprule
      \textbf{Method} & \textbf{\begin{tabular}[c]{@{}c@{}}R@4\end{tabular}} & \textbf{\begin{tabular}[c]{@{}c@{}}R@10\end{tabular}} & \textbf{\begin{tabular}[c]{@{}c@{}}R@100\end{tabular}} & \textbf{\begin{tabular}[c]{@{}c@{}}MAP@10\end{tabular}} & \textbf{\begin{tabular}[c]{@{}c@{}}MRR@10\end{tabular}} & \textbf{\begin{tabular}[c]{@{}c@{}}NDCG@10\end{tabular}} \\ \midrule
      \multicolumn{7}{c}{\textbf{Sparse (BM25)}} \\ \midrule
      Raw & 83.3 & 88.9 & 96.7 & 49.5 & 75.4 & 70.5 \\
      Qwen2.5-3B-inst (wo/qd) & 72.0 & 79.3 & 89.7 & 41.2 & 64.2 & 60.5 \\
      Qwen2.5-3B-inst (w/qd) & 75.3 & 81.2 & 89.5 & 42.7 & 65.9 & 62.4 \\
      DeepSeek-V3.1 (w/qd) & 81.1 & 86.6 & 93.3 & 49.1 & 70.6 & 66.2 \\
      vanilla RL \textit{\textcolor{gray}{(Qwen2.5-3B)}} & 82.3 & 89.9 & 95.6 & 48.8 & 77.5 & 73.2 \\
      ConvSearch-R1 \textit{\textcolor{gray}{(Qwen2.5-3B)}} & 83.0 & 90.2 & 96.0 & 51.1 & 77.0 & 72.3 \\ \midrule
      ACQO \textit{\textcolor{gray}{(ours, Qwen2.5-3B)}} & \textbf{86.9} & \textbf{91.6} & \textbf{97.5} & \textbf{51.2} & \textbf{77.7} & \textbf{74.2} \\ \midrule
      \multicolumn{7}{c}{\textbf{Dense (ANCE)}} \\ \midrule
      Raw & 68.3 & 74.8 & 86.1 & 34.8 & 60.4 & 59.5 \\
      Qwen2.5-3B-inst (wo/qd) & 64.6 & 70.7 & 81.8 & 32.8 & 56.6 & 56.0 \\
      Qwen2.5-3B-inst (w/qd) & 67.0 & 73.0 & 81.8 & 34.9 & 57.5 & 57.3 \\
      DeepSeek-V3.1 (w/qd) & 77.4 & 82.5 & 88.9 & 46.1 & 66.8 & 65.7 \\
      vanilla RL \textit{\textcolor{gray}{(Qwen2.5-3B)}} & 79.2 & 83.9 & 89.1 & 41.1 & 75.5 & 74.4 \\
      ConvSearch-R1 \textit{\textcolor{gray}{(Qwen2.5-3B)}} & 75.0 & 79.4 & 87.5 & 44.4 & 72.8 & 72.2 \\ \midrule
      ACQO \textit{\textcolor{gray}{(ours, Qwen2.5-3B)}} & \textbf{82.2} & \textbf{85.8} & \textbf{91.2} & \textbf{49.6} & \textbf{73.4} & \textbf{73.6} \\ \bottomrule
      \end{tabular}
        \caption{Retrieval performance comparison on HotpotQA (\%). (qd: query decomposition)}
    \label{tab:hotpotqa_results}
  
    \vspace{-4pt}
    \end{table}

  \subsection{Main Results}
  \label{subsec:main_results}
  
  Table~\ref{tab:topiocqa_results} and ~\ref{tab:hotpotqa_results} show the retrieval performance of our method on TopiOCQA and HotpotQA. % using BM25 and ANCE retrievers, along with comparisons to baselines.
  We further present comprehensive end-to-end RAG experiment results in \ref{subsec:e2e_qa}, which also validate the effectiveness of our method.
  
  The results on TopiOCQA demostrate that ACQO significantly outperforms most methods across different retrieval settings. Notably, our method achieves competitive performance (34.9\% MRR@3, 37.7\% NDCG@3) using self-supervised via retrieval feedback, while \textit{ConvSearch-R1} achieves a strong 37.8\% MRR@3 in sparse retrieval, this performance stems primarily from its extended reasoning process and aggressive rewrite expansion mechanisms, which are also present in other methods.
  % are amplified here to the point of consuming over 10× more tokens on average than our approach. 
  %As shown in Table~\ref{tab:gen_lengths}, its strong performance comes at the cost of over 10$\times$ more tokens than our method, making it too slow and resource-heavy for practical end-to-end RAG use,
  As shown in Table~\ref{tab:gen_lengths} and Table~\ref{tab:latency}, its strong performance comes at the cost of both over 10× more tokens than our method and a 9.1× increase in inference latency for the observed gains, making it too slow and resource-heavy for practical end-to-end RAG use,
  which gains driven by scale, not scalable design. However, ACQO demonstrates superior generalization capabilities, achieving the best R@10 (62.6\%) and R@100 (83.2\%) performance on sparse retrieval. In dense retrieval settings, ACQO shows remarkable effectiveness, attaining competitive MRR@3 (36.6\%), NDCG@3 (39.4\%) and R@10 (65.6\%), demonstrating its ability to work across different retrieval architectures.

  %On HotpotQA, ACQO achieves the best results across all metrics under both sparse and dense retrieval settings while using only a 3B parameter model. Query decomposition generally helps models perform better than their non decomposition counterparts yet even the strongest decomposition baselines including DeepSeek V3.1 fall short of the raw query in sparse retrieval, which indicates that straightforward decomposition or instruction based rewriting can harm retrieval effectiveness on this multi hop dataset. 
  % On HotpotQA, using only a 3B parameter model, ACQO achieves the best results across all metrics under both sparse and dense retrieval settings. Query decomposition generally helps models outperform their non-decomposition counterparts; yet even the strongest decomposition baselines (e.g., DeepSeek V3.1) fall short of the raw query baseline in sparse retrieval. This indicates that straightforward decomposition or instruction-based rewriting can harm retrieval effectiveness on this multi-hop dataset. In contrast ACQO avoids such degradation and significantly outperforms the raw query: in sparse retrieval, it achieves 86.9\% R@4 (+3.6\%) and 91.6\% R@10 (+2.7\%); in dense retrieval, it reaches 82.2\% R@4 (+13.9\%) and 85.8\% R@10 (+11.0\%), outperforming the best baseline by +4.8\% and +3.3\% respectively. These results demonstrate that ACQO successfully bridges the gap between query decomposition and retrieval alignment, delivering superior and robust performance without relying on larger models or sacrificing efficiency.
  
  On HotpotQA, using only a 3B parameter model, ACQO achieves the best results across all metrics under both sparse and dense retrieval settings. Notably, ACQO outperforms ConvSearch-R1 on this more challenging multi-hop dataset (49.6\% vs. 44.4\% MAP@10), demonstrating superior decomposition capability. %Query decomposition generally helps models outperform their non-decomposition counterparts; yet even the strongest decomposition baselines (e.g., DeepSeek V3.1) fall short of the raw query baseline in sparse retrieval. This indicates that straightforward decomposition or instruction-based rewriting can harm retrieval effectiveness on this multi-hop dataset.
  While query decomposition often boosts performance over non-decomposition methods, even strong baselines (e.g., DeepSeek V3.1) fall short of raw queries in sparse retrieval—indicating straightforward decomposition may harm multi-hop retrieval. In contrast, ACQO avoids such degradation and significantly outperforms the raw query: in sparse retrieval, it achieves 86.9\% R@4 (+3.6\%) and 91.6\% R@10 (+2.7\%); in dense retrieval, it reaches 82.2\% R@4 (+13.9\%) and 85.8\% R@10 (+11.0\%), outperforming the best baseline by +4.8\% and +3.3\% respectively. These results demonstrate that ACQO successfully bridges the gap between query decomposition and retrieval alignment, delivering superior and robust performance without relying on larger models or sacrificing efficiency.

  \subsection{Evaluation on Out-of-Distribution Data}
  A critical strength of our ACQO framework lies in its strong generalization to entirely unseen datasets. As shown in Table~\ref{tab:multihop_retrieval_comparison}, when evaluated on MultiHop-RAG, ACQO consistently outperforms raw queries and all baselines across different retrievers. It achieves 49.7\% R@4 compared to 45.7\% for raw, with clear gains of 4\% using \textit{llm-embedder} and 3\% using \textit{bge-large-en-v1.5}, confirming its compatibility with varying retrieval architectures. ACQO maintains strong performance on unseen domains and query types, indicating it learns domain-invariant reformulation principles. All gains are achieved zero-shot without fine-tuning, which confirming it generalizes beyond dataset-specific patterns, making it highly adaptable to real-world retrieval systems with shifting data.
  \begin{table}[t]
  \scriptsize
  \renewcommand{\arraystretch}{0.5}
  \setlength{\tabcolsep}{8pt}
  \centering
  \begin{tabular}{@{}lcccc@{}}
    \toprule
    \multirow{2}{*}{\textbf{Method}} & \multicolumn{4}{c}{\textbf{bge-large-en-v1.5}} \\ \cmidrule(l){2-5} 
     & MRR@10 & MAP@10 & R@10 & R@4 \\ \midrule
    Raw & 45.5 & 21.5 & 81.3 & 62.5 \\
    Qwen2.5-3B(w/qd) & 44.8 & 21.2 & 80.5 & 61.7 \\ \midrule
    ACQO \textit{\textcolor{gray}{(ours)}} & \textbf{47.7} & \textbf{23.6} & \textbf{84.0} & \textbf{65.5} \\ \midrule
    \multirow{2}{*}{\textbf{Method}} & \multicolumn{4}{c}{\textbf{llm-embedder}} \\ \cmidrule(l){2-5} 
     & \multicolumn{1}{l}{MRR@10} & MAP@10 & R@10 & R@4 \\ \midrule
    Raw & 32.9 & 14.4 & 65.7 & 45.7 \\
    Qwen2.5-3B(w/qd) & 33.2 & 14.7 & 65.7 & 45.9 \\ \midrule
    ACQO \textit{\textcolor{gray}{(ours)}} & \textbf{35.6} & \textbf{17.3} & \textbf{72.6} & \textbf{49.7} \\ \bottomrule
    \end{tabular}
  \caption{Retrieval performance comparison on MultiHop-RAG (\%).}
  \label{tab:multihop_retrieval_comparison}

  \end{table}
  \vspace{-5pt}
  \begin{table*}[t]
    \scriptsize
    % 第一个表格：TopiOCQA（仅保留NDCG@3、R@3、R@10）
    \centering
    \renewcommand{\arraystretch}{0.5}
    \setlength{\tabcolsep}{4pt}
    \begin{tabular}{@{}l|cccccc|cccccc@{}}
    \toprule
    \textbf{Dataset} & \multicolumn{6}{c|}{\textbf{TopiOCQA}} & \multicolumn{6}{c}{\textbf{HotpotQA}} \\ \midrule
    \textbf{Retriver} & \multicolumn{3}{c}{\textbf{Sparse}} & \multicolumn{3}{c|}{\textbf{Dense}} & \multicolumn{3}{c}{\textbf{Sparse}} & \multicolumn{3}{c}{\textbf{Dense}} \\ \midrule
    \textbf{Method} & \textbf{NDCG@3} & \textbf{R@3} & \textbf{R@10} & \textbf{NDCG@3} & \textbf{R@3} & \textbf{R@10} & \textbf{MAP@10} & \textbf{R@3} & \textbf{R@10} & \textbf{MAP@10} & \textbf{R@3} & \textbf{R@10} \\ \midrule
    - wo/ RSF & 35.0 & 42.1 & 58.8 & 38.8 & 46.6 & 63.4 & 51.2 & 83.5 & 91.1 & 49.0 & 80.1 & \textbf{85.6} \\
    - wo/ Stage II & 24.9 & 30.6 & 49.1 & 36.3 & 44.2 & 64.9 & \textbf{52.0} & 84.6 & \textbf{91.8} & 40.6 & 69.4 & 75.1 \\
    - wo/ QD & 36.5 & 44.1 & 61.1 & 38.7 & 46.1 & 63.2 & 49.9 & 83.1 & 90.5 & 42.3 & 79.6 & 84.7 \\ \midrule
    \textbf{ACQO} & \textbf{37.7} & \textbf{45.8} & \textbf{62.6} & \textbf{39.4} & \textbf{47.8} & \textbf{65.6} & 51.2 & \textbf{84.8} & 91.6 & \textbf{49.4} & \textbf{80.5} & \textbf{85.6} \\ \bottomrule
    \end{tabular}
    \caption{Ablation study on retrieval performance (\%).}
    \label{tab:ablation}
  
    \vspace{-5pt}
    \end{table*}

  \subsection{Ablation Study}
  \label{subsec:ablationstudy}
  In this work, we have presented ACQO with three core components: Query Decomposition (QD) for adaptive query optimization, Rank-Score Fusion (RSF) for robust result aggregation, and a two-stage Curriculum Reinforcement Learning approach for stable training. We conduct comprehensive ablation studies on these components across both TopiOCQA and HotpotQA datasets to understand their individual contributions.
  As shown in Table \ref{tab:ablation}, all three components are essential for optimal performance, with removing any single component leading to noticeable performance drops across both dense and sparse retrievers.
  
  \textbf{Rank-Score Fusion (RSF)} emerges as the most critical component, with its removal causing the most significant performance degradation on TopiOCQA (37.7\% → 35.0\% NDCG@3 for sparse, 39.4\% → 38.8\% for dense), demonstrating that effective aggregation of multiple query results is fundamental to our approach.
  \textbf{Curriculum Reinforcement Learning} shows dramatic impact on training stability, with substantial performance drops without it (37.7\% → 24.9\% NDCG@3 for sparse on TopiOCQA), indicating that the convergence phase is essential for stable learning. We argue that Stage I (exploration) discovers diverse query reformulation strategies, while Stage II (convergence) refines these strategies for optimal performance.
  \textbf{Query Decomposition (QD)} shows moderate but consistent improvements (37.7\% → 36.5\% NDCG@3 for sparse on TopiOCQA), which aligns with expectations since TopiOCQA primarily involves disambiguation rather than complex query decomposition, yet QD still provides benefits for handling multi-faceted information needs.
  
  The synergistic effects of all components create a robust framework where each component compensates for the limitations of others, establishing that ACQO requires all three components working in concert to achieve state-of-the-art performance.

  \subsection{Training Dynamics Analysis}
\begin{figure}[H]
    \centering
    % 第一排
      \includegraphics[width=0.47\linewidth]{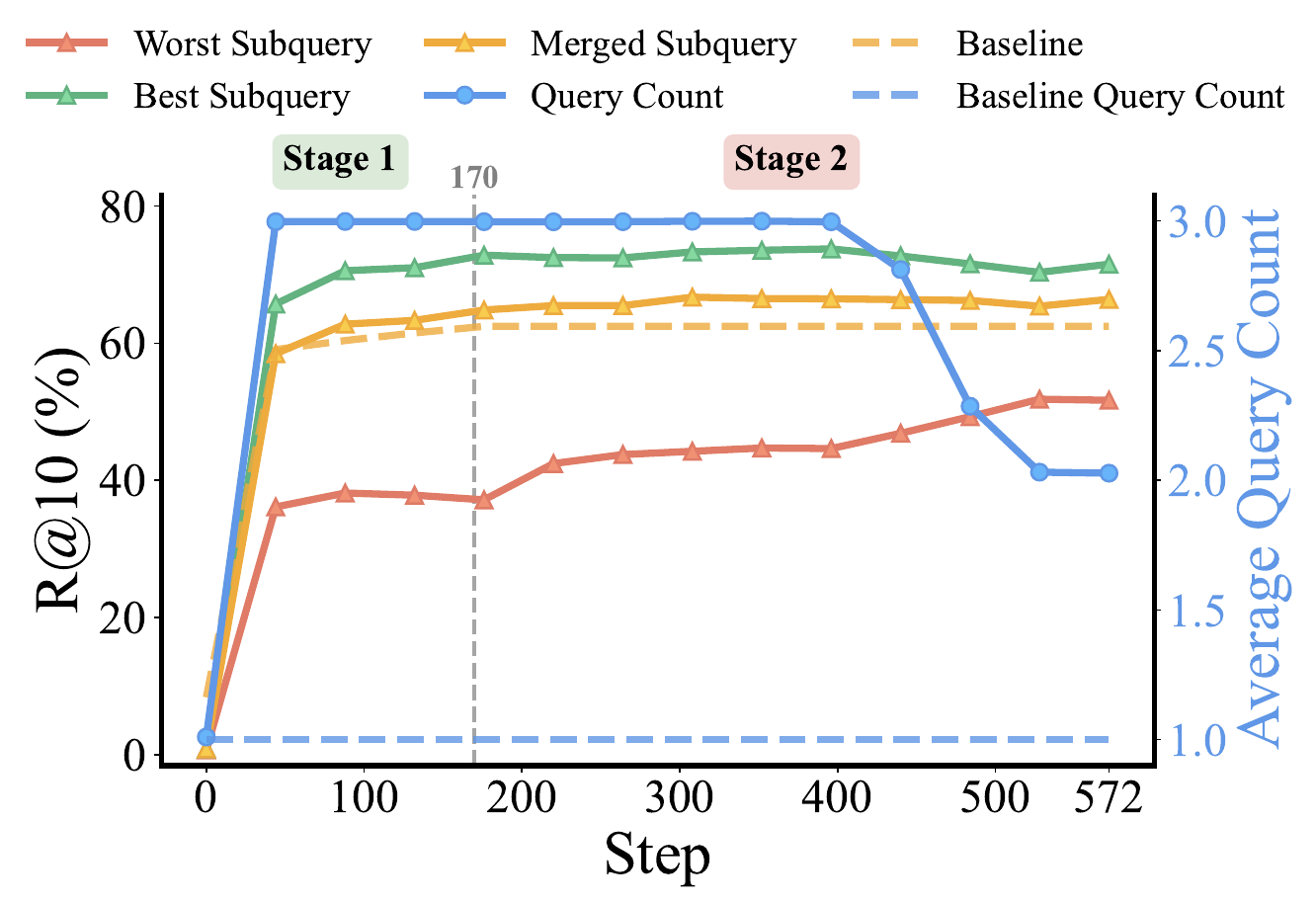}
      \includegraphics[width=0.47\linewidth]{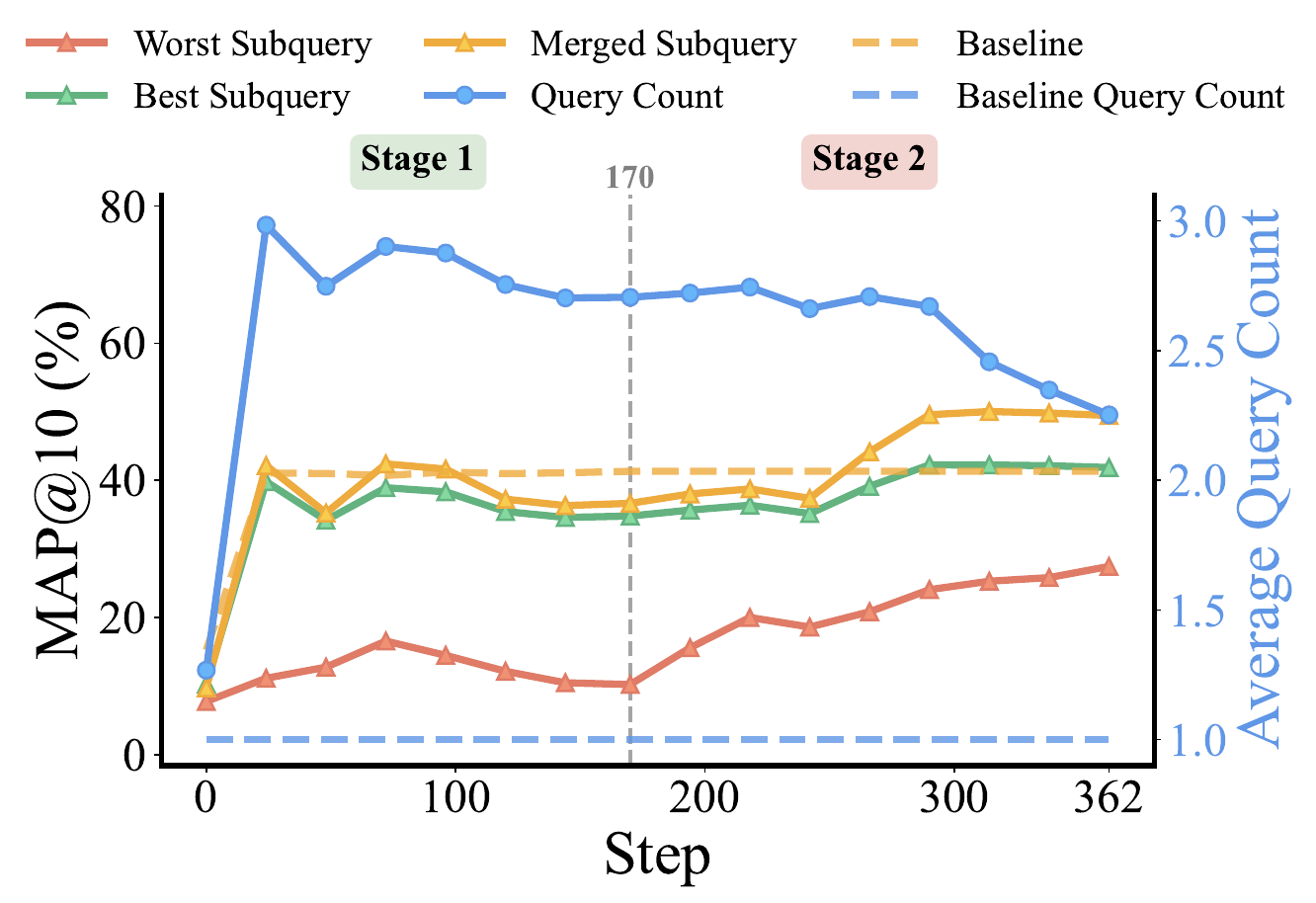}
    \caption{Query adaptation and performance improvement on TopiOCQA(L) and HotpotQA(R).}
    \label{fig:training_dynamics}
    \vspace{-4pt}
  \end{figure}

  Figure~\ref{fig:training_dynamics} illustrates the training progression of our two-stage curriculum learning approach on TopiOCQA and HotpotQA datasets. The results demonstrate the expected behavior of our adaptive query optimization framework.
  
  As shown in both datasets, the average query count follows a characteristic \textit{explore-then-converge} pattern: initially increasing during Stage I (exploration) as the model learns to decompose complex queries, then stabilizing or slightly decreasing during Stage II (convergence) as the model refines its decomposition strategies. This behavior aligns with our curriculum learning design, where the model first explores diverse query reformulation patterns before strategy convergence.
  
  The retrieval performance (R@10 for TopiOCQA, MAP@10 for HotpotQA) shows consistent improvement throughout training, with merged subqueries significantly outperforming baselines and approaching the performance of best subqueries. Notably, different retrievers result in different optimal query counts after training, which corroborates our finding that effective query optimization requires retriever-specific adaptation.
  
  The training dynamics validate that our two-stage approach successfully balances exploration and exploitation, achieving both improved retrieval effectiveness and computational efficiency through adaptive query count optimization. We demonstrate the effectiveness of ACQO through state-of-the-art performance on TopiOCQA and HotpotQA datasets. In addition, the experimental results indicate that ACQO learns retriever-specific optimization strategies, with different retrievers yielding different optimal query patterns. Furthermore, ACQO exhibits superior performance in challenging settings such as generalization on unseen datasets and computational efficiency with smaller models.
  
  \section{Conclusion}
  
  In this work, we propose ACQO, a two-stage reinforcement learning framework that addresses complex query optimization in RAG systems through self-supervised retrieval feedback, which leverages retrieval signals via adaptive query decomposition and rank-score fusion to provide retriever-specific guidance for query optimization. Experimental results demonstrate state-of-the-art performance on TopiOCQA and HotpotQA, while achieving 9.1× faster inference than strong baselines. Our analysis further reveals that ACQO learns retriever-specific optimization strategies, with each retriever yielding distinct optimal query patterns. Furthermore, our framework demonstrates superior performance in challenging scenarios, including strong generalization to unseen datasets and efficient operation with smaller models, establishing a powerful, efficient, and generalizable solution for next-generation RAG systems.

\bibliography{iclr2026_conference}

@article{huang2024survey,
  title={A survey on retrieval-augmented text generation for large language models},
  author={Huang, Yizheng and Huang, Jimmy},
  journal={arXiv preprint arXiv:2404.10981},
  year={2024}
}

@inproceedings{yu2020few,
  title={Few-shot generative conversational query rewriting},
  author={Yu, Shi and Liu, Jiahua and Yang, Jingqin and Xiong, Chenyan and Bennett, Paul and Gao, Jianfeng and Liu, Zhiyuan},
  booktitle={Proceedings of the 43rd International ACM SIGIR conference on research and development in Information Retrieval},
  pages={1933--1936},
  year={2020}
}

@inproceedings{sheng2025hybridflow,
  title={Hybridflow: A flexible and efficient rlhf framework},
  author={Sheng, Guangming and Zhang, Chi and Ye, Zilingfeng and Wu, Xibin and Zhang, Wang and Zhang, Ru and Peng, Yanghua and Lin, Haibin and Wu, Chuan},
  booktitle={Proceedings of the Twentieth European Conference on Computer Systems},
  pages={1279--1297},
  year={2025}
}

@article{ammann2025question,
  title={Question Decomposition for Retrieval-Augmented Generation},
  author={Ammann, Paul JL and Golde, Jonas and Akbik, Alan},
  journal={arXiv preprint arXiv:2507.00355},
  year={2025}
}

@article{perez2020unsupervised,
  title={Unsupervised question decomposition for question answering},
  author={Perez, Ethan and Lewis, Patrick and Yih, Wen-tau and Cho, Kyunghyun and Kiela, Douwe},
  journal={arXiv preprint arXiv:2002.09758},
  year={2020}
}

@article{liu2024ra,
  title={Ra-isf: Learning to answer and understand from retrieval augmentation via iterative self-feedback},
  author={Liu, Yanming and Peng, Xinyue and Zhang, Xuhong and Liu, Weihao and Yin, Jianwei and Cao, Jiannan and Du, Tianyu},
  journal={arXiv preprint arXiv:2403.06840},
  year={2024}
}

@inproceedings{lai2024adacqr,
  title={AdaCQR: Enhancing Query Reformulation for Conversational Search via Sparse and Dense Retrieval Alignment},
  author={Lai, Yilong and Wu, Jialong and Zhang, Congzhi and Sun, Haowen and Zhou, Deyu},
  booktitle={Proceedings of the 31st International Conference on Computational Linguistics},
  pages={7698--7720},
  year={2025}
}

@inproceedings{jang2023itercqr,
  title={IterCQR: Iterative Conversational Query Reformulation with Retrieval Guidance},
  author={Jang, Yunah and Lee, Kang-il and Bae, Hyunkyung and Lee, Hwanhee and Jung, Kyomin},
  booktitle={Proceedings of the 2024 Conference of the North American Chapter of the Association for Computational Linguistics: Human Language Technologies (Volume 1: Long Papers)},
  pages={8114--8131},
  year={2024}
}

@misc{deepseekai2024deepseekv3technicalreport,
      title={DeepSeek-V3 Technical Report}, 
      author={DeepSeek-AI},
      year={2024},
      eprint={2412.19437},
      archivePrefix={arXiv},
      primaryClass={cs.CL},
      url={https://arxiv.org/abs/2412.19437}, 
}

@misc{qwen2.5,
    title = {Qwen2.5: A Party of Foundation Models},
    url = {https://qwenlm.github.io/blog/qwen2.5/},
    author = {Team Qwen},
    month = {September},
    year = {2024}
}

@article{lewis2020retrieval,
  title={Retrieval-augmented generation for knowledge-intensive nlp tasks},
  author={Lewis, Patrick and Perez, Ethan and Piktus, Aleksandra and Petroni, Fabio and Karpukhin, Vladimir and Goyal, Naman and K{\"u}ttler, Heinrich and Lewis, Mike and Yih, Wen-tau and Rockt{\"a}schel, Tim and others},
  journal={Advances in neural information processing systems},
  volume={33},
  pages={9459--9474},
  year={2020}
}

@inproceedings{vakulenko2021comparison,
  title={A comparison of question rewriting methods for conversational passage retrieval},
  author={Vakulenko, Svitlana and Voskarides, Nikos and Tu, Zhucheng and Longpre, Shayne},
  booktitle={European Conference on Information Retrieval},
  pages={418--424},
  year={2021},
  organization={Springer}
}

@inproceedings{zhang2024adaptive,
  title={Adaptive Query Rewriting: Aligning Rewriters through Marginal Probability of Conversational Answers},
  author={Zhang, Tianhua and Li, Kun and Luo, Hongyin and Wu, Xixin and Glass, James R and Meng, Helen},
  booktitle={EMNLP},
  year={2024}
}

@inproceedings{xu2024search,
  title={Search-in-the-chain: Interactively enhancing large language models with search for knowledge-intensive tasks},
  author={Xu, Shicheng and Pang, Liang and Shen, Huawei and Cheng, Xueqi and Chua, Tat-Seng},
  booktitle={Proceedings of the ACM Web Conference 2024},
  pages={1362--1373},
  year={2024}
}

@article{feng2023synergistic,
  title={Synergistic interplay between search and large language models for information retrieval},
  author={Feng, Jiazhan and Tao, Chongyang and Geng, Xiubo and Shen, Tao and Xu, Can and Long, Guodong and Zhao, Dongyan and Jiang, Daxin},
  journal={arXiv preprint arXiv:2305.07402},
  year={2023}
}

@article{azad2019query,
  title={Query expansion techniques for information retrieval: a survey},
  author={Azad, Hiteshwar Kumar and Deepak, Akshay},
  journal={Information Processing \& Management},
  volume={56},
  number={5},
  pages={1698--1735},
  year={2019},
  publisher={Elsevier}
}

@article{wang2023query2doc,
  title={Query2doc: Query expansion with large language models},
  author={Wang, Liang and Yang, Nan and Wei, Furu},
  journal={arXiv preprint arXiv:2303.07678},
  year={2023}
}

@inproceedings{gao2023precise,
  title={Precise zero-shot dense retrieval without relevance labels},
  author={Gao, Luyu and Ma, Xueguang and Lin, Jimmy and Callan, Jamie},
  booktitle={Proceedings of the 61st Annual Meeting of the Association for Computational Linguistics (Volume 1: Long Papers)},
  pages={1762--1777},
  year={2023}
}

@article{singh2025agentic,
  title={Agentic retrieval-augmented generation: A survey on agentic rag},
  author={Singh, Aditi and Ehtesham, Abul and Kumar, Saket and Khoei, Tala Talaei},
  journal={arXiv preprint arXiv:2501.09136},
  year={2025}
}

@article{adlakha2022topiocqa,
  title={Topiocqa: Open-domain conversational question answering with topic switching},
  author={Adlakha, Vaibhav and Dhuliawala, Shehzaad and Suleman, Kaheer and de Vries, Harm and Reddy, Siva},
  journal={Transactions of the Association for Computational Linguistics},
  volume={10},
  pages={468--483},
  year={2022},
  publisher={MIT Press One Broadway, 12th Floor, Cambridge, Massachusetts 02142, USA~…}
}

@article{yang2018hotpotqa,
  title={HotpotQA: A dataset for diverse, explainable multi-hop question answering},
  author={Yang, Zhilin and Qi, Peng and Zhang, Saizheng and Bengio, Yoshua and Cohen, William W and Salakhutdinov, Ruslan and Manning, Christopher D},
  journal={arXiv preprint arXiv:1809.09600},
  year={2018}
}

@article{tang2024multihoprag,
  title={Multihop-rag: Benchmarking retrieval-augmented generation for multi-hop queries},
  author={Tang, Yixuan and Yang, Yi},
  journal={arXiv preprint arXiv:2401.15391},
  year={2024}
}

@article{bajaj2016msmarco,
  title={Ms marco: A human generated machine reading comprehension dataset},
  author={Bajaj, Payal and Campos, Daniel and Craswell, Nick and Deng, Li and Gao, Jianfeng and Liu, Xiaodong and Majumder, Rangan and McNamara, Andrew and Mitra, Bhaskar and Nguyen, Tri and others},
  journal={arXiv preprint arXiv:1611.09268},
  year={2016}
}

@article{xiong2020ance,
  title={Approximate nearest neighbor negative contrastive learning for dense text retrieval},
  author={Xiong, Lee and Xiong, Chenyan and Li, Ye and Tang, Kwok-Fung and Liu, Jialin and Bennett, Paul and Ahmed, Junaid and Overwijk, Arnold},
  journal={arXiv preprint arXiv:2007.00808},
  year={2020}
}

@article{zhang2023llm-embedder,
  title={Retrieve anything to augment large language models},
  author={Zhang, Peitian and Xiao, Shitao and Liu, Zheng and Dou, Zhicheng and Nie, Jian-Yun},
  journal={arXiv preprint arXiv:2310.07554},
  year={2023}
}

@inproceedings{xiao2024bge-large-en-v1.5,
  title={C-pack: Packed resources for general chinese embeddings},
  author={Xiao, Shitao and Liu, Zheng and Zhang, Peitian and Muennighoff, Niklas and Lian, Defu and Nie, Jian-Yun},
  booktitle={Proceedings of the 47th international ACM SIGIR conference on research and development in information retrieval},
  pages={641--649},
  year={2024}
}

@software{Liu_LlamaIndex_2022,
author = {Liu, Jerry},
doi = {10.5281/zenodo.1234},
month = {11},
title = {{LlamaIndex}},
url = {https://github.com/jerryjliu/llama_index},
year = {2022}
}

@inproceedings{sheng2025verl,
  title={Hybridflow: A flexible and efficient rlhf framework},
  author={Sheng, Guangming and Zhang, Chi and Ye, Zilingfeng and Wu, Xibin and Zhang, Wang and Zhang, Ru and Peng, Yanghua and Lin, Haibin and Wu, Chuan},
  booktitle={Proceedings of the Twentieth European Conference on Computer Systems},
  pages={1279--1297},
  year={2025}
}

@article{lin2021pyserini,
  title={Pyserini: An easy-to-use python toolkit to support replicable ir research with sparse and dense representations},
  author={Lin, Jimmy and Ma, Xueguang and Lin, Sheng-Chieh and Yang, Jheng-Hong and Pradeep, Ronak and Nogueira, Rodrigo},
  journal={arXiv preprint arXiv:2102.10073},
  year={2021}
}

@article{johnson2019faiss,
  title={Billion-scale similarity search with GPUs},
  author={Johnson, Jeff and Douze, Matthijs and J{\'e}gou, Herv{\'e}},
  journal={IEEE Transactions on Big Data},
  volume={7},
  number={3},
  pages={535--547},
  year={2019},
  publisher={IEEE}
}

@inproceedings{mao2023large,
  title={Large Language Models Know Your Contextual Search Intent: A Prompting Framework for Conversational Search},
  author={Mao, Kelong and Dou, Zhicheng and Mo, Fengran and Hou, Jiewen and Chen, Haonan and Qian, Hongjin},
  booktitle={Findings of the Association for Computational Linguistics: EMNLP 2023},
  pages={1211--1225},
  year={2023}
}

@inproceedings{mo2024chiq,
  title={CHIQ: Contextual History Enhancement for Improving Query Rewriting in Conversational Search},
  author={Mo, Fengran and Ghaddar, Abbas and Mao, Kelong and Rezagholizadeh, Mehdi and Chen, Boxing and Liu, Qun and Nie, Jian-Yun},
  booktitle={EMNLP},
  year={2024}
}

@inproceedings{yoon2025ask,
  title={Ask Optimal Questions: Aligning Large Language Models with Retriever’s Preference in Conversation},
  author={Yoon, Chanwoong and Kim, Gangwoo and Jeon, Byeongguk and Kim, Sungdong and Jo, Yohan and Kang, Jaewoo},
  booktitle={Findings of the Association for Computational Linguistics: NAACL 2025},
  pages={5899--5921},
  year={2025}
}

@article{zhu2025convsearch,
  title={ConvSearch-R1: Enhancing Query Reformulation for Conversational Search with Reasoning via Reinforcement Learning},
  author={Zhu, Changtai and Wang, Siyin and Feng, Ruijun and Song, Kai and Qiu, Xipeng},
  journal={arXiv preprint arXiv:2505.15776},
  year={2025}
}

@misc{song2024surveyqueryoptimizationlarge,
      title={A Survey of Query Optimization in Large Language Models}, 
      author={Mingyang Song and Mao Zheng},
      year={2024},
      eprint={2412.17558},
      archivePrefix={arXiv},
      primaryClass={cs.CL},
      url={https://arxiv.org/abs/2412.17558}, 
}

@article{yu2025dapo,
  title={Dapo: An open-source llm reinforcement learning system at scale},
  author={Yu, Qiying and Zhang, Zheng and Zhu, Ruofei and Yuan, Yufeng and Zuo, Xiaochen and Yue, Yu and Dai, Weinan and Fan, Tiantian and Liu, Gaohong and Liu, Lingjun and others},
  journal={arXiv preprint arXiv:2503.14476},
  year={2025}
}

@article{jiang2025deepretrieval,
  title={Deepretrieval: Hacking real search engines and retrievers with large language models via reinforcement learning},
  author={Jiang, Pengcheng and Lin, Jiacheng and Cao, Lang and Tian, Runchu and Kang, SeongKu and Wang, Zifeng and Sun, Jimeng and Han, Jiawei},
  journal={arXiv preprint arXiv:2503.00223},
  year={2025}
}

@article{thakurbeir,
  title={BEIR: A Heterogeneous Benchmark for Zero-shot Evaluation of Information Retrieval Models},
  author={Thakur, Nandan and Reimers, Nils and R{\"u}ckl{\'e}, Andreas and Srivastava, Abhishek and Gurevych, Iryna}
}
% \bibliographystyle{iclr2026_conference}

\appendix

\section{Experimental Details}
\label{App:expdetails}
\subsection{Retrieval}
In TopiOCQA and HotpotQA, we use the BM25 retriever implemented by Pyserini~\citep{lin2021pyserini}, and the ANCE retriever implemented by Faiss~\citep{johnson2019faiss}. 
The hyperparameters of BM25 are set to $k_{1}=0.9, b=0.4$ for TopiOCQA, and $k_{1}=1.2, b=0.75$ for HotpotQA during all training and evaluation. 
For ANCE, to improve training efficiency, we first generate embeddings for documents and then build an HNSW index using Faiss's \texttt{IndexHNSWFlat}, with parameters $M=64$ and $\textit{ef\_construction}=2000$. The index construction for HotpotQA partially follows~\citep{jiang2025deepretrieval}. During evaluation, we use \texttt{IndexFlatIP} to construct a flat index to ensure accuracy. 
In MultiHop-RAG, we follow its original setup with the LlamaIndex~\citep{Liu_LlamaIndex_2022} framework and adopt BGE-large-en-v1.5 and LLM-Embedder as retrievers. 
Both retrievers use a chunking strategy with \texttt{chunk\_size}=256 and \texttt{chunk\_overlap}=20, splitting the 609 original documents into 7786 chunks. 
For BGE-large-en-v1.5, we follow the official recommendation and add the instruction ''Represent this sentence for searching relevant documents:'' when converting text into embeddings.

\subsection{Training and evaluation}

\paragraph{Evaluation Metrics.}
For TopiOCQA, we employ Mean Reciprocal Rank@K (MRR@K), Normalized Discounted Cumulative Gain@K (NDCG@K), and Recall@K (R@K) as evaluation metrics. 
For HotpotQA and MultiHop-RAG, we additionally use Mean Average Precision@10 (MAP@10) for assessment. For MultiHop-RAG, we follow the evaluation code and metric provided in the benchmark.

\paragraph{Retrieval Systems.}
We evaluated the performance of model under both sparse and dense retrievers. 
For TopiOCQA and HotpotQA, we select \textbf{BM25} as the sparse retriever and ANCE as the dense retriever, where \textbf{ANCE}~\citep{xiong2020ance} is trained on MS-MARCO~\citep{bajaj2016msmarco} document retrieval tasks.
For MultiHop-RAG, we use \textbf{bge-large-en-v1.5}~\citep{xiao2024bge-large-en-v1.5} and \textbf{llm-embedder}~\citep{zhang2023llm-embedder} as the retrievers.

\paragraph{Implementation.}
We deploy Qwen2.5-3B as the backbone and train the model individually on TopiOCQA and HotpotQA, following the two-stage CRL described in \S\ref{gen_inst}. 
We use verl~\citep{sheng2025verl} as our RL training framework, and adopt DAPO~\citep{yu2025dapo} as the optimization algorithm, training the models under BM25 and ANCE retrievers independently.

\paragraph{Training Hyperparameters.}

We adopt the default hyperparameters established by ConvSearch-R1~\citep{zhu2025convsearch} and the verl framework~\citep{sheng2025hybridflow}, rather than performing dataset-specific tuning. The only modification we make is setting the maximum response length to 256 tokens (vs. 1024 in ConvSearch-R1), since ACQO generates concise sub-queries rather than chain-of-thought reasoning, reducing the required output length.

Specifically, for both TopiOCQA and HotpotQA, the models are trained under BM25 and ANCE retrievers with essentially the same hyperparameter configuration across both stages.
% \paragraph{Optimization.}
We adopt DAPO optimization with mini-batch size 64 and micro-batch size 8 per GPU (8 GPUs in total). The actor learning rate is $1\times10^{-6}$ with gradient clipping at 1.0. Entropy regularization is disabled ($\text{entropy\_coeff}=0$). KL control is not used in reward shaping ($\text{use\_kl\_in\_reward=False}$). The clipping ratios are set as $[0.2, 0.28]$ with an additional coefficient $\text{clip\_ratio\_c}=10.0$, following the default configuration of DAPO.
We sample $K=8$ rollouts per query. Generation uses temperature $0.8$, top-$p=0.8$, and top-$k=-1$ during training; for validation we set $\text{temperature}=0.7$, $\text{top-}p=0.8$, and $\text{top-}k=20$. Dynamic batch sizing is enabled for efficiency, with maximum batched tokens set to 11408 and GPU memory utilization capped at 0.8. We set the training batch size to 256 and the generation batch size to 512. 
For HotpotQA, the maximum prompt length is 512 tokens, while for TopiOCQA it is 1536 tokens. In both datasets, the maximum response length is fixed to 256 tokens. We set the decay coefficient $\eta=0.6$ for HotpotQA, while $\eta=1.0$ is used for TopiOCQA. For stage~II reward design, we set $k^*=3$ for TopiOCQA and $k^*=0$ for HotpotQA. For the training epochs, Stage~I CRL is trained for 2 epochs on TopiOCQA and 3 epochs on HotpotQA. 
Stage~II CRL is trained for 10 epochs on TopiOCQA with ANCE retriever and 8 epochs on TopiOCQA with BM25 retriever. 
For HotpotQA, Stage~II is trained for 4 epochs with ANCE and 6 epochs with BM25.

% \paragraph{Training schedule.}
% Training runs for 15 epochs on 8 GPUs (single node). We save checkpoints every 22 steps and run evaluation every 36 steps. Logging is performed via console and SwanLab. The project and experiment names are recorded in the trainer configuration.

\subsection{Datasets}

We use three datasets in our experiments: TopiOCQA, HotpotQA, and MultiHop-RAG. 
All experiments are conducted on standard training/test splits and document collections, as summarized in Table~\ref{tab:statistics}. 
For HotpotQA, we follow the corpus provided by the BEIR benchmark~\citep{thakurbeir}, which standardizes the document collection for retrieval-based evaluation. 

TopiOCQA follows CC-BY-NC-SA 4.0 license. HotpotQA follows CC-BY-SA 4.0 license. MultiHop-RAG follows ODC-BY license. All reuse, modification, and related operations on these datasets strictly adhered to the copyright statements of the data owners. 

\begin{table}[ht]
\label{tab:datasets}
\centering
\scriptsize
\setlength{\tabcolsep}{4pt}
\begin{tabular}{llccc}
\toprule
\textbf{Dataset} & \textbf{Split} & \textbf{\#Queries} & \textbf{\#documents} & \textbf{\#Golden / query} \\
\midrule
\multirow{2}{*}{TopiOCQA} 
  & train & 45450 & \multirow{2}{*}{25700592} & \multirow{2}{*}{1} \\
  & test  & 2514  &  &  \\
\midrule
\multirow{2}{*}{HotpotQA} 
  & train & 85000 & \multirow{2}{*}{5233329} & \multirow{2}{*}{2} \\
  & test  & 7405  &  &  \\
\midrule
MultiHop-RAG 
  & test  & 2556  & 7786 & multiple \\
\bottomrule
\end{tabular}
\caption{Statistics of datasets used in our experiments. 
``\#Golden / query'' denotes the number of golden documents associated with each query.}
\label{tab:statistics}
\end{table}

\paragraph{Data Collection}
Our method does not require any supervised data; instead, it employ RL  with different levels of difficulty across the two RL stages(Section~\ref{subsec:CRL}). 
In Stage~I CRL, we use the full official training set for TopiOCQA. 
For HotpotQA, however, given the large training set size and relatively high initial performance, we first filter out the higher-performing samples and retain only 50\% of the data for Stage~I training. 
In Stage~II CRL, we apply dynamic filtering with the Stage~I model (Section~\ref{subsec:stage2crl}), again retaining roughly 50\% of the samples. Basically, we set $\tau_{thres}=\frac{5}{3}$ and rollouts $n=8$.
This guides the model to focus on moderately difficult instances, thereby improving learning efficiency and convergence in Stage~II.

\section{Further Experiments}

\subsection{Case Study}

In this section, we begin with a representative case to further discuss how our method improves retrieval performance.

\begin{figure*}[h]
  \centering
  % 第一张子图
  \begin{subfigure}[t]{0.8\linewidth}
    \centering
    \includegraphics[width=\linewidth]{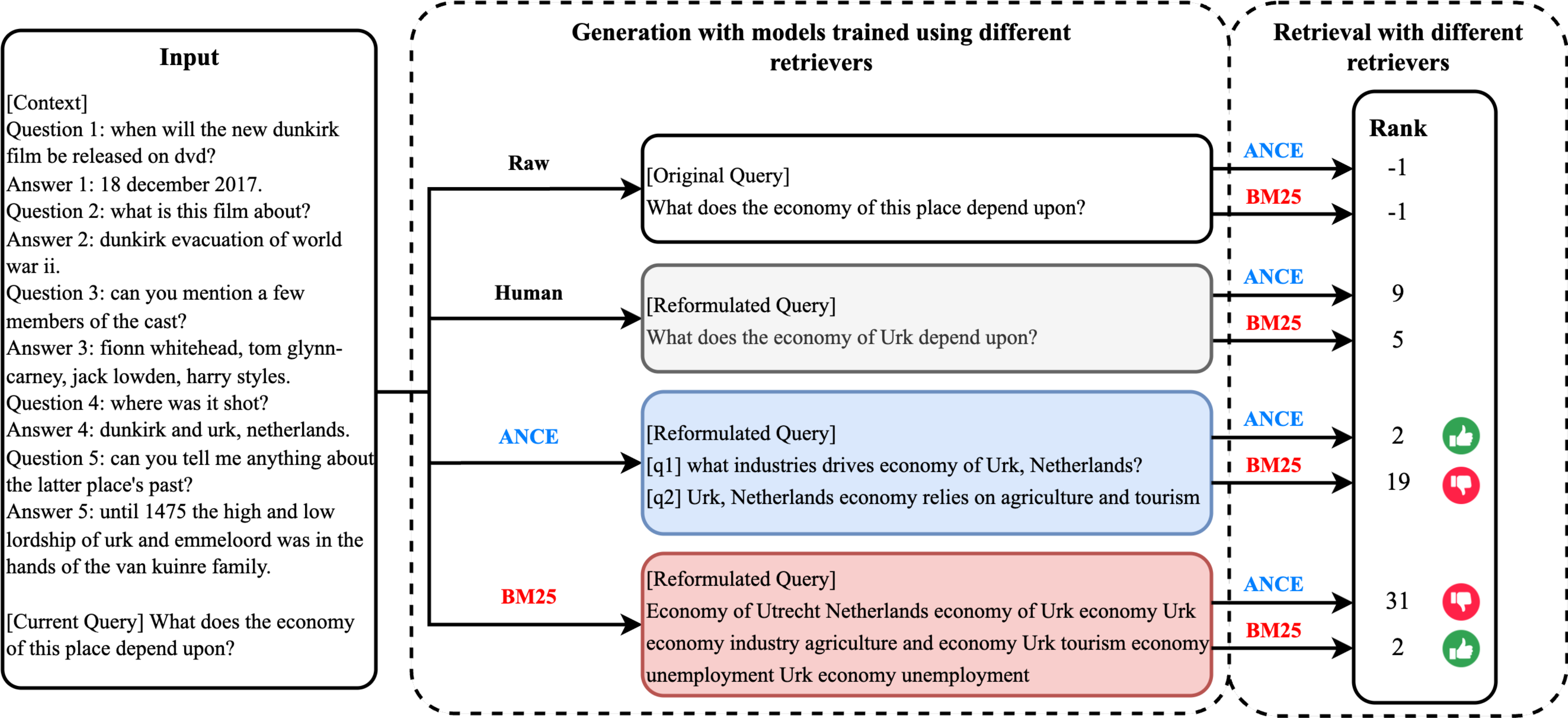}
    \caption{TopiOCQA}
    \label{fig:tpcasestudy}
  \end{subfigure}
  \hfill
  % 第二张子图
  \begin{subfigure}[t]{0.8\linewidth}
    \centering
    \includegraphics[width=\linewidth]{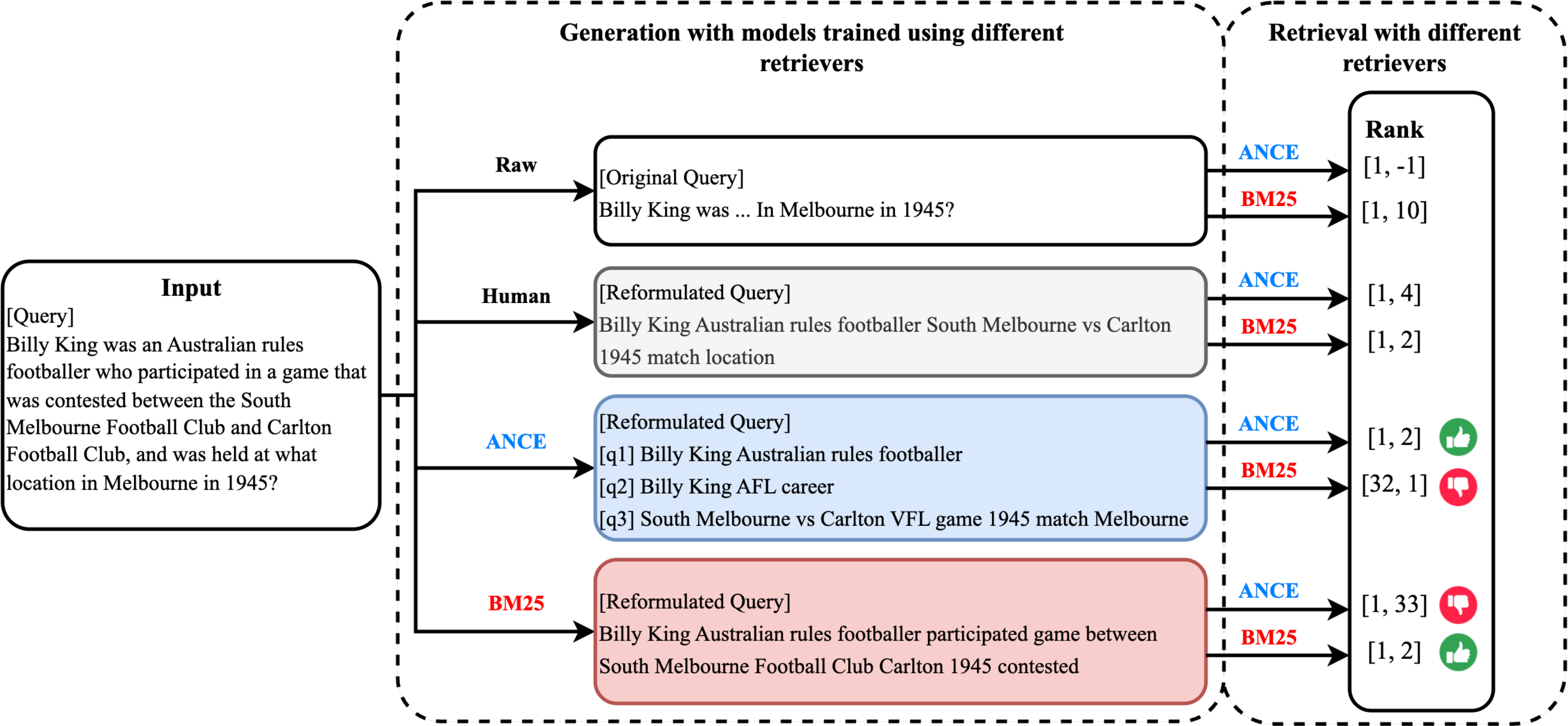}
    \caption{HotpotQA}
    \label{fig:hpcasestudy}
  \end{subfigure}
  \caption{Comparison of queries generated by models trained with different retrievers and their retrieval performance across different retrievers.}
  \label{fig:casestudy}
\end{figure*}

In Figure~\ref{fig:casestudy}, we compare the performance across different training–retrieval combinations on different datasets, i.e., the effectiveness of reformulated queries generated by models trained with a specific retriever when evaluated on other types of retrievers.

From the perspective of retrieval performance, we observe that retrieval effectiveness drops significantly when switching retrievers; moreover, model-generated reformulations outperform human-written ones (i.e., reformulations that are intuitively considered correct). This suggests that the evaluation of query quality should be retriever-dependent and may not necessarily align with human intuition. 

From the perspective of query generation behavior, queries generated with different retrievers vary in both quantity and style, indicating that the reformulation style learned by the model is closely tied to the retriever used during training.
\paragraph{What behavior does the model learn when trained with a specific retriever?} 
Models trained with the ANCE retriever tend to generate multiple queries resembling natural language questions or statements, capturing complete semantic relations and emphasizing keywords or core entities with fewer stopwords. In contrast, models trained with the BM25 retriever are inclined to generate a single query that explicitly enumerates all relevant keywords.

\paragraph{Are these behaviors aligned with retriever preferences?} 
Indeed, the observed behaviors are consistent with the characteristics favored by different retrievers. 
For dense retrievers such as ANCE, queries expressed in a natural language style, often decomposed into multiple sub-queries, better capture semantic relations and leverage embedding-based similarity. 
In contrast, sparse retrievers like BM25 prefer a single query containing exhaustive keyword coverage, where term frequency and exact lexical overlap dominate ranking. 
This alignment indicates that our model effectively adapts its reformulation strategy to the underlying retriever, learning to generate query styles that are inherently compatible with the retriever’s scoring mechanism.

\paragraph{Why does our method also yield improvements on TopiOCQA?} 
TopiOCQA consists of single-intent questions, each associated with only one golden document, which suggests that the optimal query should ideally be a single reformulation. 
Traditional approaches mainly rely on \textbf{expansion}, where the model generates a lengthy reformulation that conveys complete semantic information while leveraging its parametric knowledge to give an answer of the question, thereby increasing semantic similarity with candidate passages (see Table~\ref{tab:gen_lengths}). 
In practice, however, we observe that the model often employs \textbf{rephrasing}—for example, expressing the same intent as either a question or a declarative statement—to broaden the search space and consequently achieve better retrieval results. Its advantages lie in stronger readability and higher efficiency, while also mitigating the negative impact of erroneous expansions when the model encounters unfamiliar or ambiguous queries, thereby improving robustness.

\begin{table}[h]
\centering
\scriptsize
\setlength{\tabcolsep}{6pt}
\begin{tabular}{llccc}
\toprule
\textbf{Method} & \textbf{Retrieval} & \textbf{Reasoning} & \textbf{Response} & \textbf{Total} \\
\midrule
ConvSearch-R1   & -     & 106 & 248 & 354 \\
\midrule
\multirow{2}{*}{ACQO \textit{\textcolor{gray}{(ours)}}} 
                & ANCE  & -   & 28  & 28  \\
                & BM25  & -   & 36  & 36  \\
\bottomrule
\end{tabular}
\caption{Comparison of generation lengths across methods.}
\label{tab:gen_lengths}
\end{table}

\paragraph{Why is human-preferred query reformulation worse than model-generated?} 
Here we collectively refer to strong instruction models (e.g., DeepSeek-V3) and human-written rewrites as \emph{human-preferred query reformulation}, since such models can generate queries that are generally regarded as high-quality under instruction or few-shot settings. 
In contrast, we denote reformulations produced by our trained models as \emph{model-generated query reformulation}. 
However, experimental results show that human-preferred reformulations still underperform compared to our method or other advanced baselines. 
Based on the above analysis, we summarize two main reasons: 

(1) \textbf{Human-preferred methods do not know what constitutes a retrieval-effective query.}
They tend to follow human instructions by completing the current query with context or decomposing multi-intent queries into several sub-queries. 
Yet, when no clear sub-intent is present, they fail to decide how to \textbf{decompose}, and typically do not perform \textbf{rephrasing} or \textbf{expansion}.  

(2) \textbf{Human-preferred methods are not capable of generating retrieval-specific reformulations.}
For example, their relative performance gap to state-of-the-art baselines is larger on BM25 than on ANCE, since—as we have observed earlier—some queries with ``poorer readability'' may actually perform better under BM25.  

This finding suggests that analyzing retriever-specific data generated through ACQO can provide insights into retriever preferences, which in turn can be used to optimize prompts for large language models and improve their performance on query optimization tasks.

Taken together, our method enables the model, without any supervised data and solely using retrieval performance as the reward, to autonomously adapt to the retriever type. In doing so, the model is able to capture reformulation patterns that are more compatible with the retriever, ultimately leading to optimal reformulations.

\subsection{Scaling Capabilities}

\begin{figure*}[h]
  \centering
  % 第一排
  \begin{subfigure}[t]{0.48\linewidth}
    \centering
    \includegraphics[width=\linewidth]{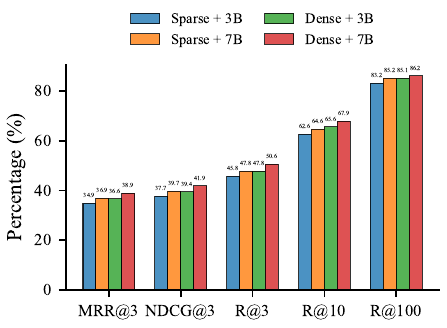}
    \caption{Comparison on TopiOCQA.}
    \label{fig:exp_a}
  \end{subfigure}
  \hfill
  \begin{subfigure}[t]{0.48\linewidth}
    \centering
    \includegraphics[width=\linewidth]{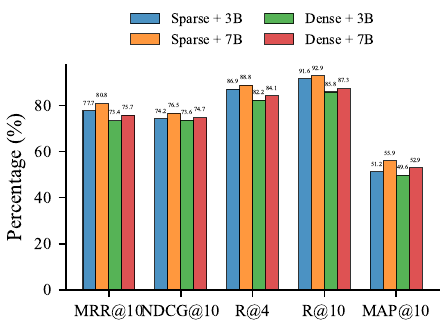}
    \caption{Comparison on HotpotQA.}
    \label{fig:exp_b}
  \end{subfigure}
  \caption{Scaling Capabilities.}
  \label{fig:scalingcapabilities}
\end{figure*}

% \paragraph{Scaling Capability.} 
Figure~\ref{fig:scalingcapabilities} presents the experimental results of our method with Qwen2.5-3B and Qwen2.5-7B.
Across both datasets and retrievers, the larger model consistently achieves better performance, demonstrating that our approach exhibits strong scaling capability.

\subsection{SFT vs. RL Comparison.}

% \paragraph{Comparison with SFT.} 
%Table~\ref{tab:sft_comparison} compares our method with SFT-based approaches (for comparisons with other baselines and ablation studies, see Sections~\ref{subsec:main_results} and~\ref{subsec:ablationstudy}). 
Table~\ref{tab:sft_comparison} presents a systematic comparison of training strategies on the TopiOCQA dataset with the ANCE retriever. For SFT baselines, the training data are constructed by rolling out the Stage~I CRL model under our framework and filtering out queries with poor rankings. These results collectively demonstrate that ACQO's two-stage curriculum reinforcement learning effectively addresses the fundamental challenges of complex query optimization, consistently outperforming both supervised baselines and vanilla RL approaches while maintaining training stability and data efficiency.
%For SFT, the training data are constructed by rollout the Stage~I CRL model under our framework and filtering out queries with poor rankings. 
%As shown in the results, both SFT on the base model and SFT on the Stage~I CRL model perform worse than our method, highlighting the effectiveness of reinforcement learning over pure supervised fine-tuning in this task.

\begin{table}[h]

\scriptsize
\setlength{\tabcolsep}{6pt}
\begin{tabular}{llccccc}
\toprule
\textbf{Method} & \textbf{MRR@3} & \textbf{NDCG@3} & \textbf{R@3} & \textbf{R@10} & \textbf{R@100} \\
\midrule
SFT & 28.4 & 30.7 & 37.3 & 53.4 & 71.5 \\
Vanilla RL  & 34.5 & 38.3 & 34.6 & 62.1 & 81.1 \\
SFT + RL & 33.4 & 37.8 & 45.7 & 61.6 & 82.2 \\
Stage I only  & 33.6 & 36.6 & 44.2 & 64.9 & \textbf{85.8} \\
Stage I + SFT  & 28.5 & 30.8 & 37.7 & 53.3 & 70.2 \\
\midrule
ACQO \textit{\textcolor{gray}{(ours)}} & \textbf{36.6} & \textbf{39.4} & \textbf{47.8} & \textbf{65.6} & 85.1 \\
\bottomrule
\end{tabular}
\caption{Comparison between SFT and RL methods on the TopiOCQA dataset with the ANCE retriever.}
\label{tab:sft_comparison}

\end{table}

\subsection{End-to-End Question Answering Evaluation}
\label{subsec:e2e_qa}

While the retrieval metrics presented above demonstrate ACQO's effectiveness in query optimization, a critical question remains: do these retrieval improvements translate to better final answers in real-world RAG applications? To address this concern, we conduct comprehensive end-to-end question answering experiments that evaluate the complete RAG pipeline from query optimization to answer generation.

\paragraph{Experimental Setup.} 
We use Qwen2.5-7B-Instruct~\citep{qwen2.5} as the reader model and DeepSeek-R1~\citep{deepseekai2024deepseekv3technicalreport} as the evaluation judge to assess answer quality on HotpotQA with the ANCE retriever. For each method (Raw Query, ConvSearch-R1, ACQO), we retrieve the top-10 documents using the optimized queries and provide them as context to the reader model, then evaluate the generated answers based on accuracy.

\paragraph{Results and Analysis.}
Table~\ref{tab:e2e_qa} presents the end-to-end evaluation results, comparing retrieval performance (MAP@10) with answer accuracy ($ACC_L$). The results reveal a strong correlation between retrieval quality and final answer accuracy across all methods. Starting from the raw query baseline (34.8\% MAP@10, 16.4\% $ACC_L$), ConvSearch-R1 achieves substantial improvements (44.4\% MAP@10, 27.7\% $ACC_L$), while ACQO further advances the state-of-the-art to 49.6\% MAP@10 and 31.6\% $ACC_L$.

\begin{table}[h]

\footnotesize
\setlength{\tabcolsep}{8pt}
\begin{tabular}{llcccc}
\toprule
\textbf{Method} & \textbf{MAP@10} & \textbf{$ACC_L$} & \textbf{$\Delta ACC_L$}  \\
\midrule
Raw Query & 34.8\% & 16.4\% & -  \\
ConvSearch-R1 & 44.4\% & 27.7\% & +11.3\%  \\
\midrule
ACQO \textit{\textcolor{gray}{(ours)}} & \textbf{49.6\%} & \textbf{31.6\%} & \textbf{+15.2\%}  \\
\bottomrule
\end{tabular}
\caption{End-to-end question answering evaluation on HotpotQA-ANCE. MAP@10 measures retrieval quality, while $ACC_L$ evaluates final answer accuracy judged by DeepSeek-R1.}
\label{tab:e2e_qa}

\end{table}

Notably, ACQO achieves a +3.9\% improvement in $ACC_L$ over ConvSearch-R1, confirming that our curriculum reinforcement learning design effectively addresses the convergence challenges in mixed-complexity query optimization. This validates that the adaptive query decomposition and robust rank-score fusion mechanism not only improve retrieval metrics but also enhance the quality of final generated answers. Moreover, ACQO reaches 9.1× lower latency, representing a favorable efficiency-accuracy trade-off for production deployment.

\subsection{Latency and Cost Analysis}
\label{subsec:latency_cost}

\paragraph{Inference Latency Analysis.}
Table~\ref{tab:latency} presents a detailed breakdown of inference latency across different pipeline stages. Our measurements on TopiOCQA-ANCE with a single H20 GPU show that ACQO adds only 31ms overhead compared to the SFT baseline (355ms vs. 324ms%, a +8.7\% increase
). More importantly, ACQO is 9.16$\times$ faster than ConvSearch-R1 (355ms vs. 3255ms) while maintaining comparable accuracy (as shown in Tables~\ref{tab:topiocqa_results} and~\ref{tab:hotpotqa_results}). This substantial speedup makes ACQO a Pareto-optimal choice for production deployment, offering the best balance between accuracy and efficiency.

The latency breakdown reveals that the additional overhead primarily comes from query generation (+23ms) and retrieval (+3ms), with our lightweight Rank-Score Fusion module contributing only 5ms. This validates our design philosophy of achieving strong performance through algorithmic innovations rather than computationally expensive components.

\begin{table}[t]
\centering
% \footnotesize
\small
\begin{minipage}[t]{\linewidth}
\centering
\scriptsize
\setlength{\tabcolsep}{5pt}
\begin{tabular}{l@{\hspace{4pt}}r@{\hspace{4pt}}r@{\hspace{4pt}}r@{\hspace{4pt}}r@{\hspace{4pt}}r@{\hspace{4pt}}r}
\toprule
\textbf{Method} & \textbf{\#Q} & \textbf{Gen} & \textbf{Retri} & \textbf{Rerank} & \textbf{Tot} & \textbf{Speed} \\
\midrule
SFT (Qwen2.5-3B) & 2514 & 297 & 27 & 0 & 324 & 1.09× \\
ACQO & 2514 & 320 & 30 & 5 & \textbf{355} & 1.0× \\
ConvSearch-R1 & 2514 & 3230 & 25 & 0 & 3255 & \textbf{0.11×} \\
\bottomrule
\end{tabular}
\vspace{0.05cm}

{\scriptsize \textit{9.16× faster than ConvSearch-R1; +31ms for +8.2\% MRR@3 vs. SFT}}
\subcaption{Avg Inference Latency (ms,TopiOCQA-ANCE)}
\label{tab:latency}

\end{minipage}
\hfill
\begin{minipage}[t]{\linewidth}
\centering
\scriptsize
\setlength{\tabcolsep}{8pt}
\begin{tabular}{lrrr}
\toprule
\textbf{Method} & \textbf{GPU-H} & \textbf{Conv} & \textbf{MAP@10} \\
\midrule
Vanilla RL & 8.4 & No & 41.1 \\
SFT + RL & 15.4 & yes & 45.3 \\
\textbf{ACQO (Stage I)} & 4.2 & yes & 42.3 \\
\textbf{ACQO (Full)} & 12.1 & yes & 49.6 \\
\bottomrule
\end{tabular}
\subcaption{Training Cost (HotpotQA-ANCE)}
\label{tab:training_cost}

\end{minipage}

\caption{Efficiency analysis: inference latency and training cost.}
\label{tab:efficiency}

\end{table}

\paragraph{Training Cost Analysis.}
Table~\ref{tab:training_cost} compares training costs on HotpotQA-ANCE using 8 H20 GPUs. Full ACQO training requires 12.1 GPU-hours, comparable to the SFT+RL baseline (15.4 GPU-hours) but without requiring any supervised query rewriting data. Notably, ACQO-Stage I converges in only 4.2 GPU-hours while achieving 42.3\% MAP@10, demonstrating efficient initial exploration.

While vanilla RL appears faster (8.4 GPU-hours), it fails to converge properly, getting stuck at a low performance ceiling (41.1\% MAP@10) due to training instability. The root cause is insufficient valid samples—the DAPO algorithm fails to collect enough qualified samples within its sampling budget (\texttt{max\_num\_gen\_batches}=20), causing premature termination with suboptimal performance. This validates the necessity of our curriculum learning strategy for stable convergence.

These results demonstrate that ACQO achieves superior performance with practical computational costs: (1) Inference efficiency: 9.16× faster than ConvSearch-R1 with minimal overhead over SFT; (2) Training efficiency: comparable cost to SFT+RL but without supervised data requirements; (3) Training stability: successful convergence where vanilla RL fails. This favorable efficiency-accuracy trade-off establishes ACQO as a practical solution for production RAG systems.

\section{Prompts}

\begin{figure*}[h]
  \centering
  \begin{subfigure}{0.99\linewidth}
    \centering
    \includegraphics[width=\linewidth]{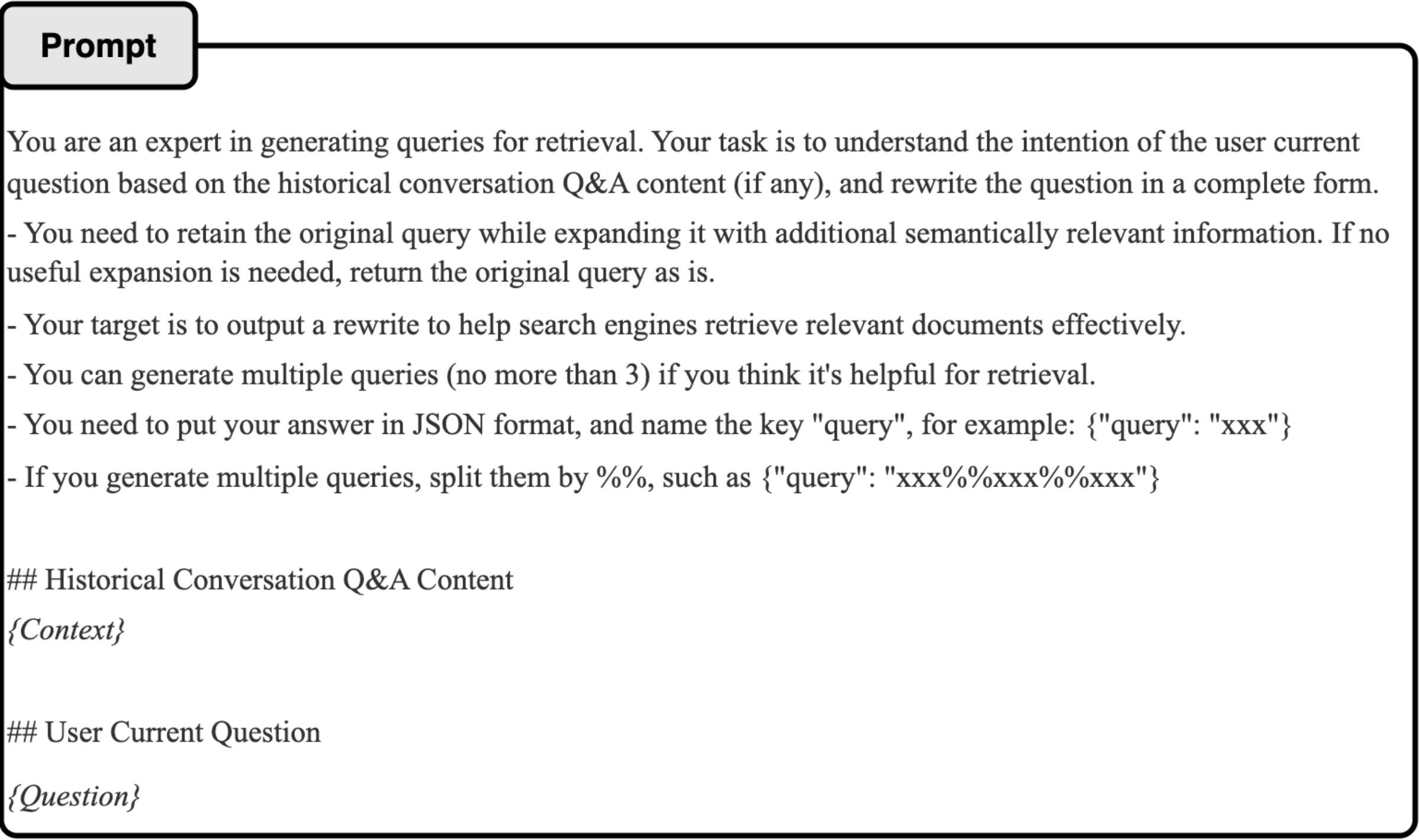}
    \caption{Prompt for standard ACQO.}
    \label{fig:prompt_std}
  \end{subfigure}
  \begin{subfigure}{0.99\linewidth}
    \centering
    \includegraphics[width=\linewidth]{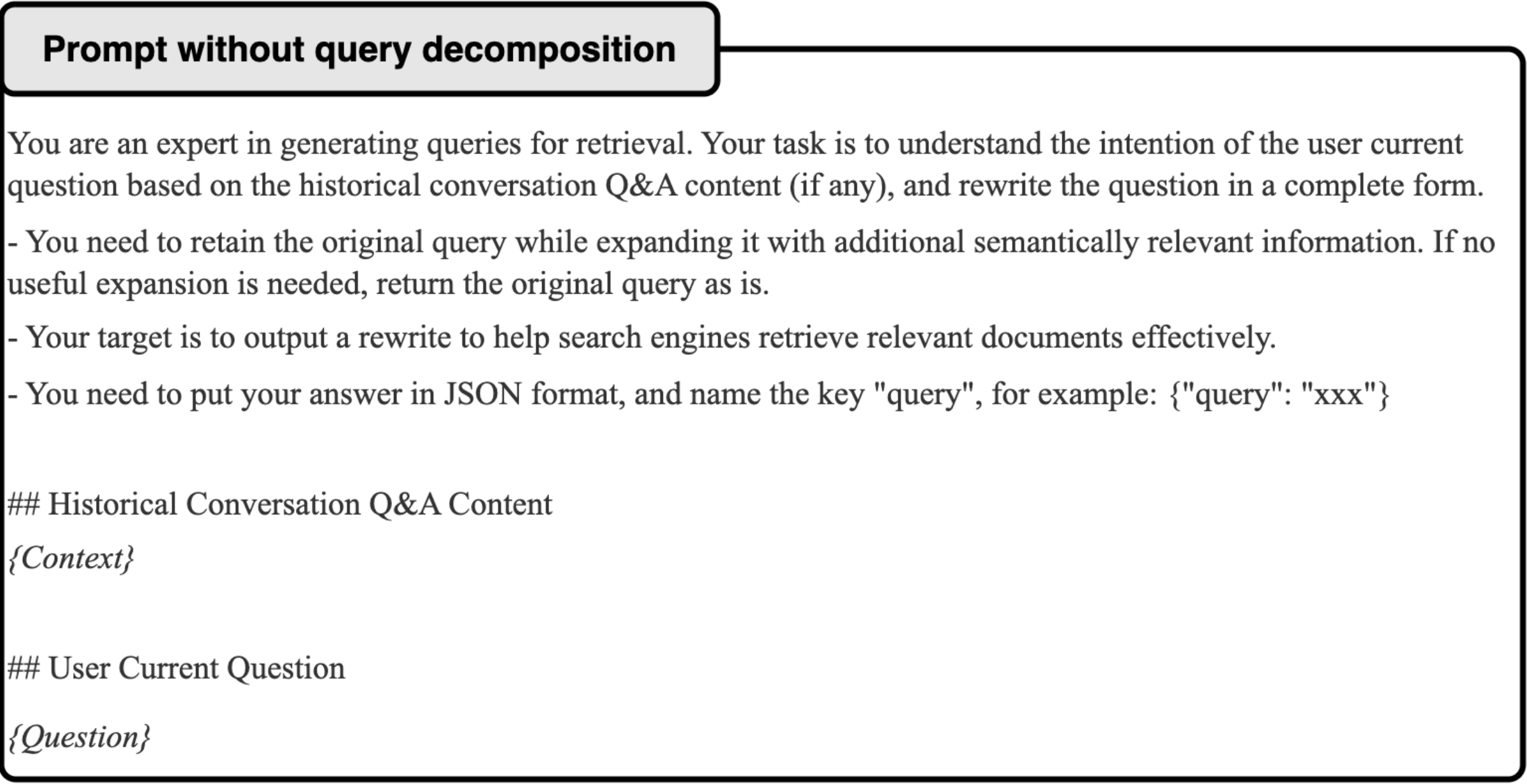}
    \caption{Prompt without query decomposition.}
    \label{fig:prompt_woqd}
  \end{subfigure}
  
  \caption{Prompts used in our experiments.}
  \label{fig:prompts}
\end{figure*}

Figure~\ref{fig:prompt_std} shows the prompt used in ACQO, which remains the same across retrievers and datasets. If no context is available, it is set to empty. The same prompt is also employed in other experiments (e.g., ablation studies and supervised fine-tuning) with query decomposition. We also provide in Figure~\ref{fig:prompt_woqd} the prompt version without query decomposition, which is used in experiments without query decomposition.

\end{document}